\pdfoutput=1
\documentclass[11pt]{article}

\usepackage[preprint]{acl}

\usepackage{times}
\usepackage{latexsym}

\usepackage{algorithm}
\usepackage{dsfont}
\usepackage{algorithmic}
\usepackage{amsmath}
\usepackage{xcolor} 
\usepackage{enumitem}
\usepackage[most]{tcolorbox} % 
\usepackage[T1]{fontenc}
\usepackage[utf8]{inputenc}

\usepackage{microtype}

\usepackage{inconsolata}
\usepackage{booktabs}
\usepackage{graphicx}
\usepackage{pgfplots}
\pgfplotsset{compat=1.18}
\usepackage{tikz}
\usepackage{pgfplots}
\usetikzlibrary{pgfplots.groupplots}
\usepackage[demo]{tikzpeople}
\usepackage{array}
\usepackage{arydshln}
\usepackage{amssymb}
\newcommand{\norm}[1]{\left\lVert#1\right\rVert}

\title{Debating for Better Reasoning: An Unsupervised Multimodal Approach}

\author{Ashutosh Adhikari \\
    School Of Informatics \\
  University of Edinburgh\\
  Edinburgh, UK \\ 
  \texttt{ashutosh.adhikari@ed.ac.uk} \\\And
  Mirella Lapata \\
  School Of Informatics \\
  University of Edinburgh\\
  Edinburgh, UK \\ 
  \texttt{mlap@ed.ac.uk} \\}

\begin{document}
\maketitle
\begin{abstract}
As Large Language Models (LLMs) gain expertise across diverse domains and modalities, scalable oversight becomes increasingly challenging, particularly when their capabilities may surpass human evaluators. Debate has emerged as a promising mechanism for enabling such oversight. In this work, we extend the debate paradigm to a multimodal setting, exploring its potential for weaker models to supervise and enhance the performance of stronger models. We focus on visual question answering (VQA), where two ``sighted'' expert vision-language models debate an answer, while a ``blind'' (text-only) judge adjudicates based solely on the quality of the arguments.  In our framework, the experts defend only answers aligned with their beliefs, thereby obviating the need for explicit role-playing and concentrating the debate on instances of expert disagreement. Experiments on several multimodal tasks demonstrate that the debate framework consistently outperforms individual expert models. Moreover, judgments from weaker LLMs can help instill reasoning capabilities in vision-language models through finetuning.\footnote{We will release the code necessary to reproduce our experiments and analysis.} 
\end{abstract}

% \section{Introduction}
\section{Introduction}
% \myworries{Introduce a problem first. Scalable oversight was proposed as an AI Safety mech. In text ppl have shown whether it can be used to ellicit truth, (2) we say can it not only predict truth but allow for an unsupervised way for collecting reasoning data-- (1) we do all this in multimodal setting}
Current approaches to aligning large language models rely heavily on human-labeled data \cite{ouyang2022traininglanguagemodelsfollow, christianorlhf2017}. However, as models gain expertise across different domains and modalities \cite{openai2024gpt4technicalreport, llama3herdmodels, deepseekai2025deepseekr1incentivizingreasoningcapability}, gathering high-quality data for training them and further aligning them becomes progressively expensive and difficult. 
% As large language models (LLMs) gain expertise across domains, . Earlier methods to align pre-trained LLMs rely heavily on human-labelled data which is ever so harder to when it comes to alignment for expertise \cite{llama3herdmodels, ouyang2022traininglanguagemodelsfollow, openai2024gpt4technicalreport}.
Previous work \cite{bowman2022SO, irving2018aisafetydebate} has investigated interactions between models as a means of 
achieving \emph{scalable oversight}. This paradigm seeks to develop techniques for supervising increasingly capable models, even in scenarios where their expertise may surpass that of human evaluators. 

%possibility of their expertise overshadowing that of humans \cite{irving2018aisafetydebate, christiano2018supervisingstronglearnersamplifying}, cotra.    
Debate  has emerged as a promising framework for enabling scalable oversight \cite{irving2018aisafetydebate, bowman2022SO, khan, deepmindSO,khan}. In a debate match, two or more expert models argue to convince a non-expert model or human judge. For instance, in a question answering setting, a question is shown to both models, they state their answers, and take turns making their statements. The judge sees the debate and decides which agent wins. Other frameworks  \cite{estronellDeb, du2023improvingfactualityreasoninglanguage,pang-etal-2022-quality} explore debate within the context of multi-agent collaboration. The core idea is to engage  models in multiple rounds of discussion aiming to converge on a \emph{consensus} response which is meant to be better than the output of a single model. \citet{subramaniam2025multiagent} propose a variant  that leverages consensus among multiple instances of the \emph{same} model, aiming to collect diverse data for further fine-tuning.  
% \myworries{enter more examples, like Askell and MAD where models come to a conclusion}. 
%as a mechanism to elicit answers with better quality by utilizing a non-expert judge to adjudicate the correct answer .  explore debating between two models to answer complex questions from the QuALITY dataset \cite{pang-etal-2022-quality}. Consequently there are works using multi-agent debate (MAD) to reach an answer by consensus lead to answers with improved quality \cite{estronellDeb, du2023improvingfactualityreasoninglanguage}. 

\begin{figure*}[t]
    \centering
 \begin{tabular}{l@{}c@{}l}
\includegraphics[width=2.95in]{boat_umbrella.png} & 
\raisebox{2.7cm}[0pt]{\begin{tcolorbox}[width=6cm,colframe=white, colback=green!30]
\small 
The image depicts a scene involving a group of people in a boat traveling along a body of water, possibly a river or canal. The boat is mainly white with a red outer deck and appears to be $\dots$ 
\end{tcolorbox}}
& \raisebox{3.5cm}[0pt]{\begin{tikzpicture}[scale=.5]
  \coordinate (head-center) at (0,0);
  \coordinate (top) at ([yshift=-2mm]head-center);
  \coordinate (left) at ([yshift=-10mm,xshift=-7mm]head-center);
  \coordinate (right) at ([yshift=-10mm,xshift=7mm]head-center);
  \draw[rounded corners=1.5mm,fill=green!30] 
  (top) to [out=-10,in=100] 
  (right) to [bend left=15]
  (left) to [out=80,in=190]
  (top);
  \draw[fill=green!30,opacity=1] (head-center) circle (0.35);
\end{tikzpicture}} \\ 
& \raisebox{0.4cm}[0pt]{\begin{tcolorbox}[width=6cm,colframe=white, colback=blue!30]
\small The image depicts a small, red and white boat on a body of water, likely a canal or river, as it steers through a narrow path surrounded by brick walls and greenery. The boat is occupied by several passengers who are standing$\dots$\vspace{-.1cm}
\end{tcolorbox}}
 & 
\raisebox{1cm}[0pt]{ 
\begin{tikzpicture}[scale=.5]
  \coordinate (head-center) at (0,0);
  \coordinate (top) at ([yshift=-2mm]head-center);
  \coordinate (left) at ([yshift=-10mm,xshift=-7mm]head-center);
  \coordinate (right) at ([yshift=-10mm,xshift=7mm]head-center);
  \draw[rounded corners=1.5mm,fill=blue!30] 
  (top) to [out=-10,in=100] 
  (right) to [bend left=15]
  (left) to [out=80,in=190]
  (top);
  \draw[fill=blue!30,opacity=1] (head-center) circle (0.35);
\end{tikzpicture}}\\
\multicolumn{3}{@{}c}{\raisebox{-0.2cm}[0pt]{\hspace*{-1.8cm}\begin{tcolorbox}[width=14cm]
\small
    \centering \textbf{Question:} Is the blue umbrella under
the black umbrella?
Please answer yes or no.
\end{tcolorbox}}}\\\\
\multicolumn{3}{@{}c}{\raisebox{-4.8cm}[0pt]{\includegraphics[width=3.2in]{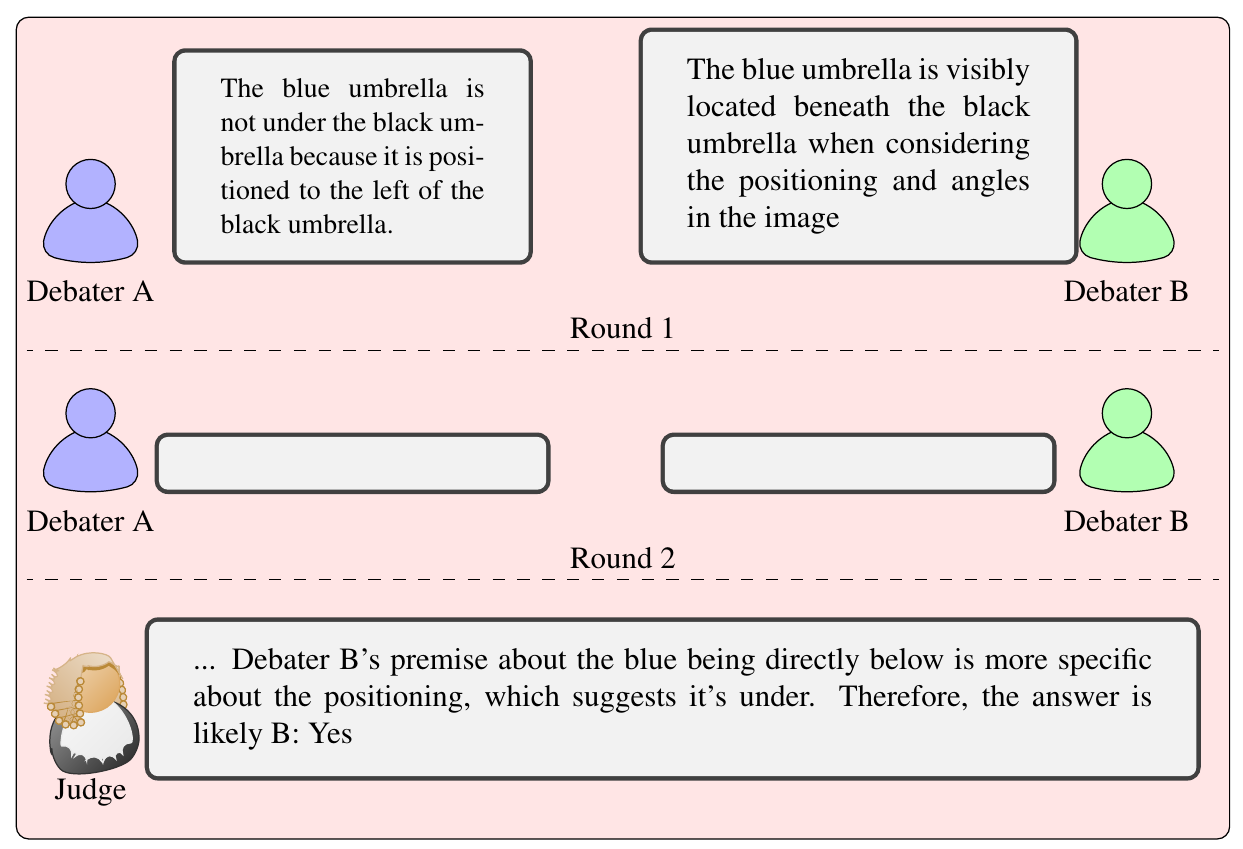}} \raisebox{-5.95cm}[0pt]{\hspace{-.47em}\includegraphics[width=3.2in]{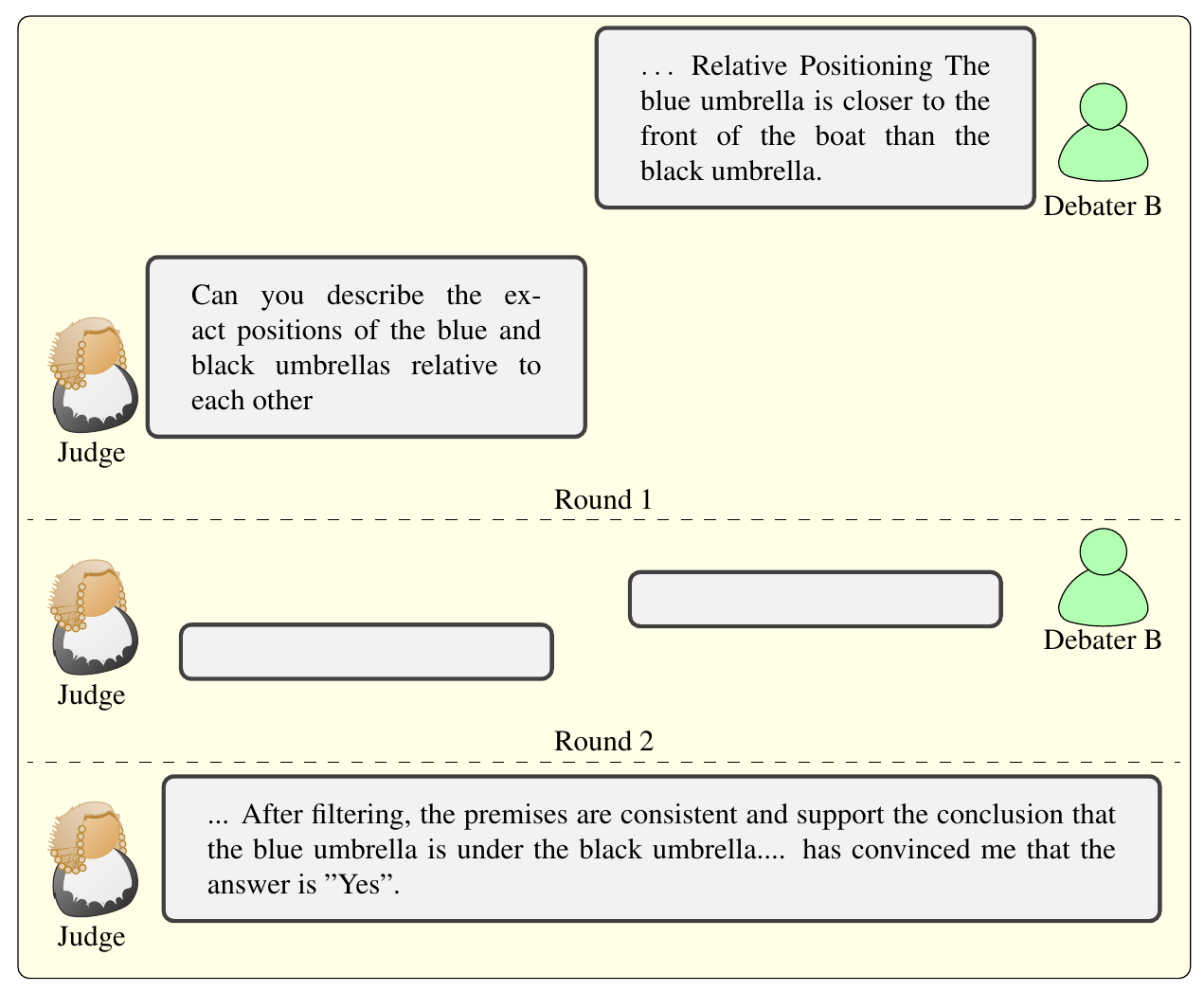}}}\\
\end{tabular}
\vspace*{-1.2cm}
    
    \caption {The visual question answering task (top) and\\ examples of debate and consultancy (bottom). In both \\protocols, the judge does not have access to the image, but only to  transcript of the debate and image descriptions  generated by the expert models (shown next to the image). Different experts and their descriptions are color-coded.}
    \label{fig:our_setup} % Label for referring to the image
\end{figure*}
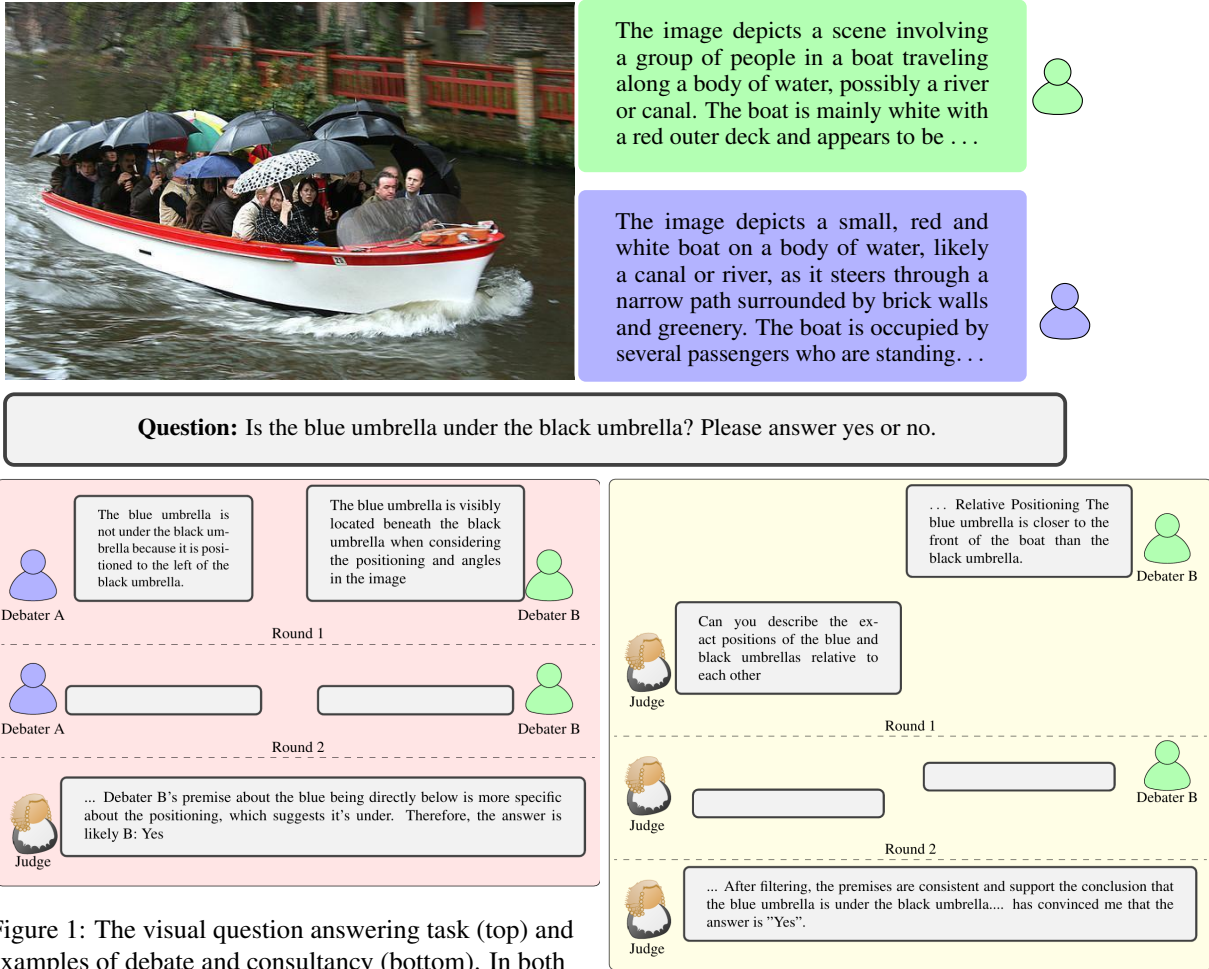

In this work, we explore the potential of debate in a multimodal setting,  as a mechanism  for
weaker models to supervise and enhance the performance of expert models. We focus on visual question answering tasks (VQA; \citealt{antol2015vqa}) where a model endowed with visual perception is meant to answer a question about an image. Following the original debate paradigm \cite{irving2018aisafetydebate}, we assume two expert  models debate the answer to a question considering  the provided visual context, while a non-expert judge determines the winner (see Figure~\ref{fig:our_setup}). While the experts are ``sighted'' and capable of simultaneously understanding both visual and linguistic modalities, our judge is ``blind'' without access to the image, and thus meant to adjudicate  based on the quality of the arguments. 

%In our knowledge this is the first comprehensive work exploring debating in a multi-modal setting. We model the non-expertise of the judge by not providing them the necessary information, the image, to answer the question on their own. This is inspired from \citet{khan, irving2018aisafetydebate} where the judge does not have access to the passages, while adjudicating the debate between two experts. In their work however, models can expose relevant parts of the passage to the judge necessary to answer the question \cite{khan, deepmindSO}.

A common framing of debate protocols has been to assign experts an answer  which they are required to defend, regardless of whether they ``believe'' it be true or correct \cite{deepmindSO, khan, du2023improvingfactualityreasoninglanguage, estronellDeb}. 
In our formulation, experts only defend answers that align with their beliefs. This obviates the need for role-playing on behalf of the experts and allows  us to debate exclusively on samples where two experts disagree.  
For judging the debates, we draw inspiration from prior work on assessing the \emph{quality} of arguments \cite{wachsmuth-etal-2017-computational, wachsmuth-werner-2020-intrinsic, el-baff-etal-2024-improving}, considering various dimensions such as their consistency, credibility, and relevance. 
%that allows a blind judge to supervise expert vision models \cite{wachsmuth-etal-2017-computational, wachsmuth-werner-2020-intrinsic, el-baff-etal-2024-improving}. 
We further use the judgments to instill reasoning capabilities in vision-language models, e.g.,~through fine-tuning the experts on the verdicts delivered by the blind judge. 

 We present experiments with open-source models across  a wide range of multimodal tasks that collectively require diverse vision-grounded skills.  Our results demonstrate that the debate framework consistently outperforms the expert models involved   in the debate and simpler scalable oversight protocols such as \emph{consultancy} \cite{khan} where a single expert model interacts with a weaker judge. 
 Our finetuning experiments further underscore the utility of debate as a mechanism for enhancing model performance, demonstrating improvements in out-of-domain settings on tasks unseen during training.
In summary our contributions are three-fold:
\begin{itemize}[noitemsep]
    \item To the best of our knowledge, we are the first to investigate  debate in multi-modal settings, and its potential for enabling weaker models to supervise stronger ones. 
    \item Drawing from argumentation research, we adapt the debate paradigm to allow models to  defend their beliefs as opposed to engaging in role-playing.
    \item Our work is the first  to employ  judgments from weaker LLMs to align expert models across modalities, without explicit human supervision (e.g., in the form of labeled data). 
\end{itemize}

\section{Related Work}
\label{sec:relwork}
A substantial body of work has previously explored debate as a framework for eliciting better quality responses from text-only language models. In their pioneering work,  \citet{irving2018aisafetydebate} motivate debate as a protocol of \textit{scalable oversight} that can be employed to supervise expert models using weaker models. As expert models are tasked to argue about an answer, a weaker judge decides which side is more convincing, even if they don't fully understand the domain or task. 
\citet{bowman2022SO} test the feasibility of  scalable oversight through interactions between humans and non-reliable LLMs, whose task is to answer difficult multiple-choice questions  \cite{hendrycks2021measuringmassivemultitasklanguage}. 

Building on this line of work, \citet{khan} and \citet{deepmindSO} investigate debate using much more capable language models on question-answering tasks involving long-form texts \cite{pang-etal-2022-quality}. They introduce the notion of a \emph{weak} judge, a language model with limited access to information that must rely solely on the experts’ arguments for determining the final answer.
The  two experts --- potentially different instantiations of the same model ---  are assigned opposing answers to debate against one another.

A distinct but related direction is the Multi-Agent Debate (MAD) framework proposed in  \citet{du2023improvingfactualityreasoninglanguage} where  multiple agents argue with one another over $n$~rounds to come to a consensus. Their results show that MAD leads to answers with improved factuality and reasoning. \citet{estronellDeb} extend this work to incentivize novelty in the debaters' arguments to prevent MAD from defaulting to majority voting. In follow-on work, \citet{estornell2025acccollab} further train debaters to improve their collaborative problem solving skills. \citet{subramaniam2025multiagent}  
 introduce a multi-agent fine-tuning approach where several models, initialized from the same base model, are independently specialized on distinct subsets of data generated from debates to encourage diversity and refinement across responses. In contrast to previous work, which primarily focuses on purely linguistic tasks, we operate in the multimodal domain,  using debate as a mechanism for a weaker model to update the beliefs of expert models. The data obtained through this process  can be also used  to finetune models that have \emph{not} participated in the debates.    

Research on computational argumentation
faces the problem of how to automatically
assess the quality of an argument (e.g.,~is it persuasive or reasonable, are its premises valid?). \citet{wachsmuth-etal-2017-computational} present  a  holistic taxonomy of all major quality dimensions for assessing natural language argumentation. \citet{gurcke-etal-2021-assessing} develop a computational model that can assess whether an argument is sufficient (i.e.,~whether an argument's premises provide enough support to justify its conclusion). We draw on this work to inform the criteria used by our weak judge to assess the quality of arguments raised during a debate.

\section{Multimodal Debate}

In this section we introduce our debate protocol for multimodal visual question answering (VQA; \citealt{antol2015vqa}) which has emerged as a key task for evaluating the ability of
vision-language models to understand images. VQA models aim  to accurately respond to questions about various 
aspects of visual content ranging from fine-grained perception to mathematical reasoning, and optical character recognition. In this work we aim to explore debating protocols for eliciting truthful answers from expert models using non-expert LLMs in a multi-modal setting, and further to utilize the arguments of the debate as training data for instilling reasoning in vision-language models. 

\paragraph{Disagreement Sets} Given a set of expert models, we first determine the samples over which they  shall debate. 
% In this work, we want to test the protocols of scalable oversight to update the beliefs of models as a mechanism to update the ``beliefs'' of the models to only debate for opinions that they believe in.
As opposed to previous work  \citep{khan, deepmindSO}, where either (or both) of the debaters are assigned an answer to defend, irrespective of their  beliefs, our debaters only argue for  answers they  believe to be true. The previous setting essentially boils down to role-playing  and does not scale easily when there are more candidate answers to a question than just two options \citep{yue2023mmmu, lu2024mathvista}. As shown in Algorithm~\ref{alg:disagreement},  for a given dataset~$\mathcal{D}$ we  determine all  samples $x \in \mathcal{D}$ over which a pair $(\mathcal{M}_i, \mathcal{M}_j)_{i \neq j}$ from our set of models $\mathbf{M}$ disagree, and~$x$ is a tuple $(q, \mathcal{I})$ consisting of a question $q$ and  image~$\mathcal{I}$. Subsequently, we run debate and consultancy matches (see definition below) only on these sets.

\begin{algorithm}[t]
\small
\caption{Disagreement Set Extraction Algorithm for Dataset $\mathcal{D}$}\label{alg:disagreement}
\begin{algorithmic}[1]
\STATE $\text{DS} \gets \emptyset$ \COMMENT{disagreement set for $\mathcal{D}$}
\STATE $\mathbf{M} \gets \text{set of models}$
\FOR{$\mathcal{M}_i \in \mathbf{M}$}
    \FOR{$\mathcal{M}_j \in \mathbf{M} \setminus \{\mathcal{M}_i\}$}
        \FOR{$x \in \mathcal{D}$}
            \IF{$\text{y}(\mathcal{M}_i({x})) \neq \text{y}(\mathcal{M}_j(x))$}
                    \STATE $\text{disagreement} \gets \{x, \mathcal{M}_i, \mathcal{M}_j\}$
                    \STATE $\mathcal{DS} \gets \mathcal{DS} \cup \{\text{disagreement}\}$
                \ENDIF
            \ENDFOR
        \ENDFOR
    \STATE $\mathbf{M} \gets \mathbf{M} \setminus \{\mathcal{M}_i\}$
    % \STATE $\text{Models}$ \gets \text{Models} \setminus $m_1$
    \ENDFOR    
    \RETURN $\mathcal{DS}$
\end{algorithmic}
\end{algorithm}
\paragraph{Debate} We illustrate our debate protocol in Figure~\ref{fig:our_setup} (bottom left). Given an image and question about its content, let $\mathcal{M}_i$ and $\mathcal{M}_j$ denote two \emph{expert} vision-language models that disagree with each other, i.e.,~they provide different answers~$y(\mathcal{M}( x))$. The two models debate, having access to the question~$q$,  image $\mathcal{I}$, and the guidelines of the debate. After completing $n$~rounds, a \emph{non-expert} judge~$\mathcal{J}$ with access to the transcript~$t_n$, adjudicates the debate to seek the ``correct'' answer. The transcript~$t_n$ is a list $_{\forall 1 \leq k \leq n}[(r_{ki}, r_{kj})]$ of responses $r$ from \emph{experts} $\mathcal{M}_i$, and $\mathcal{M}_j$ following $n$~rounds. To generate responses for the $k^{th}$ round where $k \leq n$, an expert $\mathcal{M}_i$ has access to the input and   responses from previous rounds from both the experts. As a result, $r_k = \mathcal{M}_i(t_{k-1}, x)$. We use a text-only model as our judge $\mathcal{J}$ to mimic its  lack of expertise for vision-intensive tasks. $\mathcal{J}$  provides judgement $\psi_x = \mathcal{J}(t_n, q, d_i, d_j)$ where $d_i$ and $d_j$ are image descriptions of the image $\mathcal{I}$ in sample $x$ from experts $\mathcal{M}_i$ and $\mathcal{M}_j$ respectively (see top right in Figure~\ref{sec:exp:setup}). We denote the judge's prediction (i.e.,~answer to the question) as $y_\mathcal{J} = y(\psi)$.

Our debate protocol is akin to earlier work \cite{khan, deepmindSO}, in which two text-based expert models debate the answer to a question about a story, while a separate judge makes a final decision solely based on the transcript of their debate (without access to the story). 
However, it is important to note that in their long-form question answering task, the experts were able to expose parts of relevant text to the judge using a citing mechanism. In our case, the blind judge validates the arguments provided by each expert based on detailed \emph{image descriptions}~$d$ from both experts  and the transcript of the debate.

% \paragraph{Assessing arguments for Judgement} We prompt the experts and the blind judge to engage in the debate and analyze the arguments put forward with guidelines inspired from argumentation theory \citep{wachsmuth-werner-2020-intrinsic,stab-gurevych-2017-recognizing,stahl-etal-2024-school,gurcke-etal-2021-assessing}. 
% Specifically, our prompts follow well-known criteria for assessing the quality of arguments based on consistency (an argument is internally consistent if it does not contradict itself), relevance (an argument fulfills the relevance citerion if all of its premises count in favor
% of the truth or falsity of the claim) and logical sufficiency (an argument complies with the sufficiency criterion if its premises provide
% enough evidence for accepting or rejecting
% the claim). The prompts for the judge and the debaters can be found in the Appendix \ref{sec:appendix}.

\paragraph{Consultancy} In consultancy, a single expert model $\mathcal{M}$ (the consultant) attempts to convince a non-expert judge $\mathcal{J}$ of the answer the expert believes to be true. 
% Between the rounds, the non-expert judge asks clarification questions which the expert answers in the subsequent rounds.
As in \citet{khan}, we adopt an \emph{interactive} consultancy protocol where the consultant engages in a dialogue with the judge, aiming to convince them their answer is correct by presenting supporting arguments (see bottom right in Figure~\ref{fig:our_setup}). 
The judge acts as a critic and asks the consultant probing questions.  As in debate, consultancy runs for a number of~$n$ rounds and the judge does not have access to the image itself, only the expert's description.  
At the end of each round, the blind judge decides on the right answer, based on the question, the image description and the arguments made by the consultant. 

\paragraph{Assessing the Quality of Arguments} We prompt the experts and the blind judge to engage in the debate and analyze the arguments put forward with guidelines inspired from argumentation theory \citep{wachsmuth-werner-2020-intrinsic,stab-gurevych-2017-recognizing,stahl-etal-2024-school,gurcke-etal-2021-assessing}. Specifically, we prompt expert~$\mathcal{M}_i \in \mathbf{M}$ to ground their arguments in the image~$\mathcal{I}$ and  provide response~$r_k$ for the $k^{th}$ round based on  premises that are {relevant} and {acceptable} given  question~$q$.  For the blind judge to rule in favor of an answer, we prompt the \emph{non-expert} model to follow well-known criteria for assessing the quality of arguments based on \emph{consistency} (an argument is internally consistent if it does not contradict itself), \emph{relevance} (an argument fulfills the relevance citerion if all of its premises count in favor
of the truth or falsity of the claim), and \emph{logical sufficiency} (an argument complies with the sufficiency criterion if its premises provide
enough evidence for accepting or rejecting
the claim). The prompts for the judge and the debaters can be found in  Appendix \ref{app:prompts}.

\paragraph{Extracting Reasoning Traces from Debates} We hypothesize that the rationales~$\psi$ 
 generated by the judge model to support its verdict $y_\mathcal{J}$ provide meaningful image-grounded reasoning traces. In both the consultancy and debate protocols, the judge considers arguments from the experts and verifies their credibility against the image descriptions to consolidate a list of premises that can explain one of the candidate answers. We posit that these reasoning traces $\mathbf{\chi}$ can be used to train vision-language models, enhancing their reasoning capabilities---which is a goal of \textit{scalable oversight}. %Training on judgements to align models has not been explored before. 
It is important to note that reasoning traces $\mathbf{\chi}$ are collected (through these protocols) without any explicit supervision and can be used to instill reasoning capabilities in vision-language models that have none (or even improve existing capabilities if the judge's accuracy is consistently better than either of the models). We create training data by combining the question $q$ and image $\mathcal{I}$ along with the reasoning trace $\mathcal{\chi}$ and answer obtained from the judge (see Appendix~\ref{sec:ex:output} for an example). Let~$\chi = \mathcal{E}(\psi, q)$ denote reasoning traces extracted from judgments~$\psi$ for question~$q$.
%, as a result $\chi = \mathcal{E}(\psi, q)$, where $\mathcal{E}$ extracts reasoning traces from the judgments $\psi$. 
Our training data contains tuples $(q, \mathcal{I}; \chi)$, we train expert models to generate $\chi$ from inputs $(q, \mathcal{I}) \in \mathcal{DS}$.
%\textbf{add more formal detail here, are you training a model on the disagreements only? Read all of this section and add formal detail.}

\paragraph{Rules of the Match}
For a fair comparison, we fix the number of rounds~${n}$ to be the same for consultancy and debate. Additionally, we run consultancy matches on the same samples as debate, i.e.,~where a pair of models disagree. In both scenarios, we do not assume the labels $\hat{y}$ or model accuracy are known to the judge before interacting with the experts. 
We evaluate each protocol based on the accuracy of the answer the judge selects after deliberating over the transcript of the debate. For consultancy, we report the mean judge accuracy after running consultancy matches for each model over a disagreement set.
%\paragraph{Win Rates for Debate and Consultancy}
An expert $\mathcal{M}_i$ wins a debate if they convince the judge their answer is correct. The expert's  \emph{win rate}~$\omega$ is the proportion of times they win over a set of disagreements:

{
\small
\vspace{-.8em}
\begin{align*}
\omega_i(\mathcal{M}_i, \mathcal{M}_j, \mathcal{J}) = \frac{\sum_{x \in \mathcal{DS}} \mathds{1} (y(\psi_x) = y(\mathcal{M}_i(x)))}{\norm{\mathcal{DS}}}
\end{align*}
}

\section{Experimental Setup}
\label{sec:exp:setup}

\paragraph{Datasets} We evaluate the protocols discussed in the previous section on  three datasets representative of  a variety of multimodal skills: MME \citep{fu2023mme}, MMMU \citep{yue2023mmmu}, and MathVista \citep{lu2024mathvista}. 
MME contains \textit{Yes/No} questions designed to assess    vision-language models on a  diverse set (14~in total) of perception and cognition tasks. MathVista evaluates models for visual understanding and compositional reasoning. It  is derived from~28 existing multimodal datasets evaluating math solving capabilities in vision-language models \cite{lu2024mathvista}.
MMMU \cite{yue2023mmmu}
is a benchmark created to evaluate multimodal models on complex tasks that require college-level subject knowledge across multiple disciplines and deliberate reasoning. We evaluate models  on the validation sets released by these benchmarks as the respective test sets are not publicly available. To simplify evaluation in our experiments, we only report results on questions requiring non-free-form answers from MathVista and MMMU.  Examples from these datasets can be found in Appendix~\ref{app:examples}. 

\paragraph{Models} We conduct experiments using four open-source vision-language models of comparable size, serving as debaters and consultants. The models include Molmo-7B-D-0924, Molmo-7B-O-0924 \cite{molmo2024}, LLaVA-OneVision-7B \cite{li2024llava}, and Qwen2.5-VL-7B-Instruct \cite{qwen2.5-VL}. These models are from the same weight class (7B~parameters) but with complementary skills due to  differences in their training regimes and data. For example, Molmo models are descriptive and provide evidence to the  judge through their pointing mechanism, whereas  Qwen2.5-VL  demonstrates stronger reasoning abilities. For judging debate and consultancy matches, we use a language-only model, Qwen32B \cite{qwen2.5, qwen2} that is distilled from the DeepSeek-R1 family \cite{deepseekai2025deepseekr1incentivizingreasoningcapability}, which we found  to be reliable at following instructions.% for judging the debates and consultancy.
% \textbf{add references for these models and say why you selected those.}

\paragraph{Debate and Consultancy Matches} We run two rounds of debate and consultancy, with  similar instructions. The number of rounds is limited by the context length capacity of the models participating in the debate. As mentioned earlier,  we base the prompts for the judges and  experts on argumentation research. We encourage models to ground their premises in the image, and form logical arguments to drive their points. For multiple-choice questions, the judge has the option not to select any of the answers advocated by the debaters and consultants. Our prompts are shown in Appendix~\ref{app:prompts}.

\paragraph{Finetuning on Reasoning Traces}  We extract reasoning traces from the judge's verdict for all samples in the disagreement sets. We also use Qwen32B to retrieve the traces from the transcript. We provide the prompt in Appendix~\ref{app:prompts}. For these experiments, we finetune with LoRA \cite{llora:2021} LLaVA-1.5-7B, a model which is not in our set of experts and  LLaVA-OneVision-7B, a model included in our set of experts. 
We evaluate the fine-tuned models in an out-of-domain setting, where we leave out one dataset and train on the other two (see Appendix~\ref{app:training} for training hyperparameters and Appendix~\ref{sec:disagreements} for dataset statistics). 

% We test the finetuning mechanism on all three datasets where the training data is from remaining datasets. 

%We utilize simple LoRA-based finetuning whether the reasoning from judgements from a weaker LLM can be used to improve experts that lack reasoning capabilities. We test the supervised models in an out-of-domain setting where the training data for finetuning is not obtained from the dataset we test on. 
\section{Results}
\label{sec:Results}

\begin{figure}[t]
  \centering
\begin{tikzpicture}[scale=.7]
\begin{axis}[
    xtick={0, 1, 2, 3},
    ytick={0, 1, 2, 3},
    xticklabels={Molmo-O, Molmo-D, LLaVA-1V, Qwen2.5-VL},
    yticklabels={Molmo-O, Molmo-D, LLaVA-1V, Qwen2.5-VL},
    x tick label style={rotate=45, anchor=east, font=\small},
    yticklabel style={align=right, font=\small},
    axis on top,
    colorbar,
    point meta min=0,
    point meta max=300,
    colormap/cool,
    nodes near coords,
    every node near coord/.append style={font=\small, color=black},
    enlargelimits=false,
    width=8cm,
height=8cm,
   clip mode=individual,
   axis equal image
]

\addplot [
    matrix plot,
    mesh/cols=4, % <-- important: tells pgfplots it's a 4x4 matrix
    point meta=explicit,
    draw=gray,
    line width=0.1pt,
] table [meta=data] {
x y data
0 0 0
1 0 312
2 0 303
3 0 326
0 1 312
1 1 0
2 1 234
3 1 239
0 2 303
1 2 234
2 2 0
3 2 216
0 3 326
1 3 239
2 3 216
3 3 0
};

\end{axis}
\end{tikzpicture}
\vspace{-.3cm}
  \caption{Heatmap showing disagreements between models on the MathVista dataset.}
  \label{fig:mv_disagreements}
\end{figure}
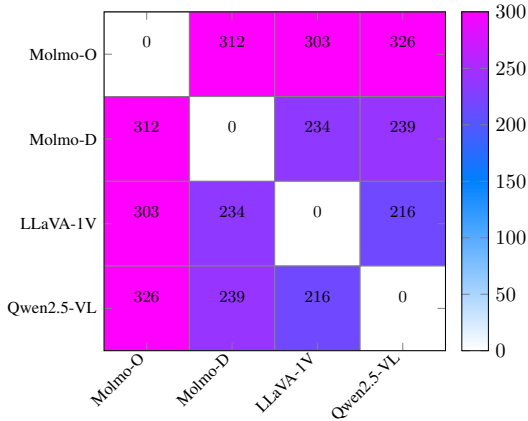

\begin{figure*}[t]
    \centering
    \includegraphics[clip,trim=4.5cm 0cm 6.5cm 0cm, width=\textwidth]{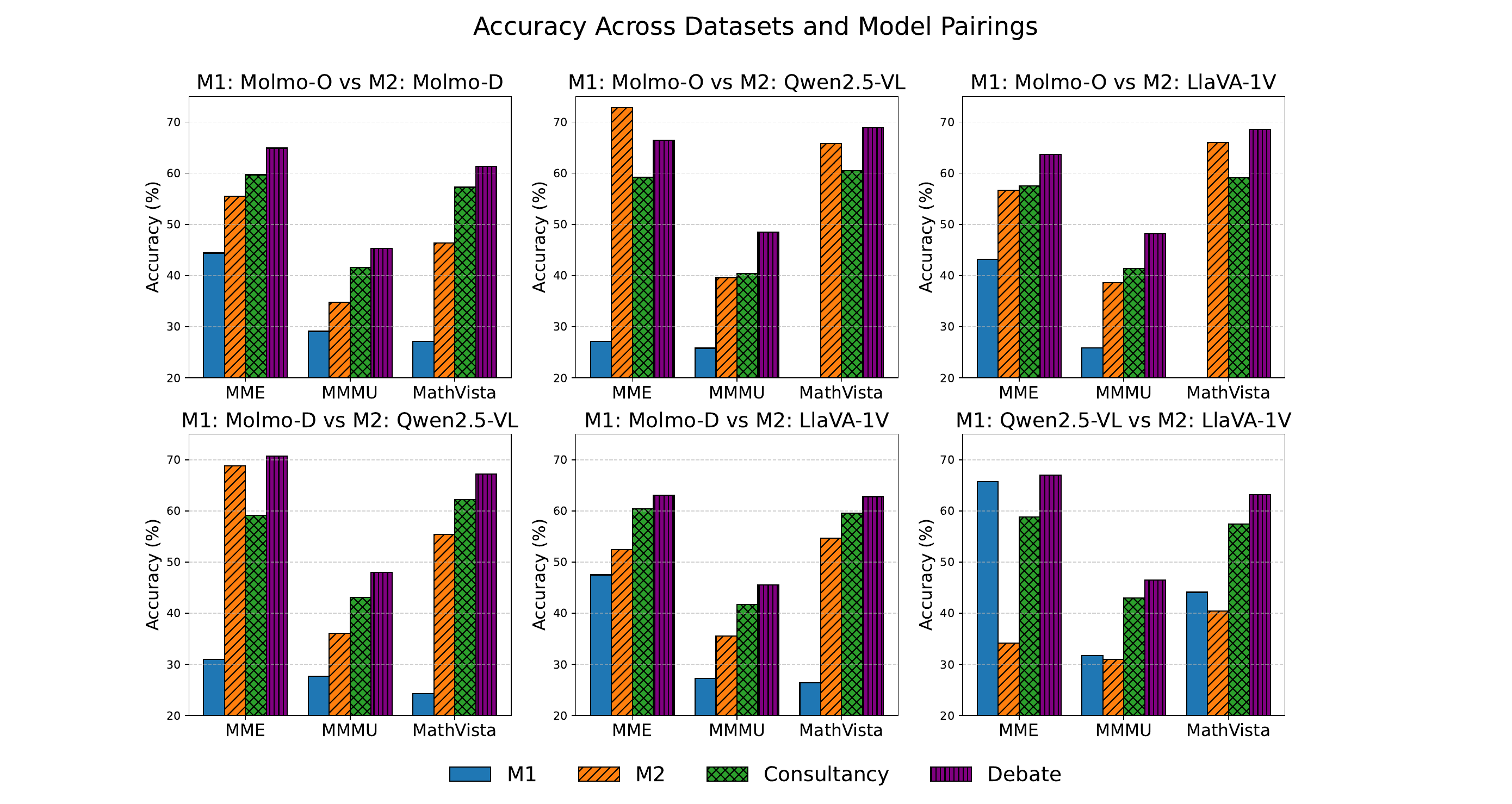} % replace with your image file name
   \vspace{-.8cm}
    \caption{Expert accuracy on disagreement sets for MME, MMMU, and MathVista datasets. Specific model pairings are shown on top of every sub-plot. 
    %The pairing on the top of every sub-plot represents the expert names. M$_{1}$ and M$_{2}$ simply represent a pair's accuracy on the set of their disagreements.
    % O, D, L and Q are shorthands for Molmo-O, Molmo-D, LlaVA-1V and Qwen2.5-VL respectively. 
    }
    \label{fig:judge_results} % Label for referring to the image
\end{figure*}

\paragraph{Model Performance and Disagreements}
Prior to  reporting our results with debate and consultancy, we evaluate expert performance on our three datasets, and also obtain disagreements for all pairs. Table~\ref{tab:full_results} summarizes the performance of our four experts. As can be seen, Qwen2.5-VL \cite{qwen2.5-VL, Qwen2VL} consistently outperforms other models, with Llava-1V trailing behind. Figure~\ref{fig:mv_disagreements} shows pairwise disagreements for MathVista for every pair of models in our experiments (see Appendix~\ref{sec:disagreements} for disagreements on the other datasets). We find that models with lower accuracy debate more with other models, whereas models with high accuracy  tend to disagree less with ane another. As a result, Molmo-O has the highest number of disagreements with all other models, and across tasks, we find Qwen-2.5-VL and LlaVA-1V disagree the least with one another. 

% Contrary to our expectation, Molmo-O-7B-0924 and Molmo-7B-D-0924 . We find that models from different family disagree more with each other. Furthermore, models that are less accurate disagree more accurate models. Whereas two more accurate models shall disagree less with one another. In MMMU, where the model performance is low for the 
% \myworries{insert the image}

\begin{figure*}[t]
    \centering
    \includegraphics[clip,trim=3.5cm 0cm 0.25cm 0cm,width=\textwidth]{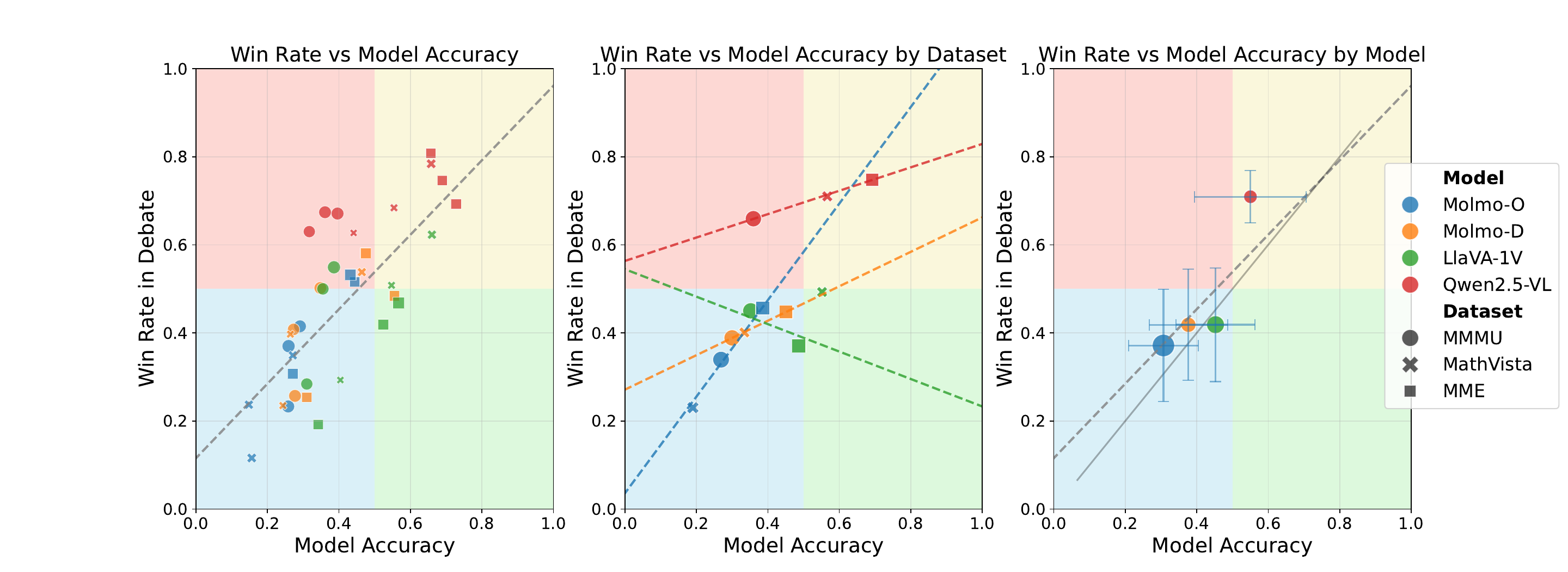}\\ % replace with your image file name
  %  \caption{\textbf{Win rates vs Model Accuracy in Debate:}. \textbf{Left} plot shows win ratio to expert accuracy for all the disagreement sets across all the datasets.  \textbf{Center} plot shows win ratio to expert accuracy by models and datasets. \textbf{Right} plot shows win ratio to expert accuracy by models. The solid line in the right plot is $y = x$. In all the subplots, the dotted lines are fit linearly based on the size of the disagreement sets. The quadrants are marked to highlight whether a model is misleading.}
    \label{fig:win_acc_deb} % Label for referring to the image
%[clip,trim=4cm 0cm 2.5cm 0cm, width=\textwidth]
%    \centering
    \includegraphics[clip,trim=3.5cm 0cm 0.25cm 0cm,width=\textwidth]{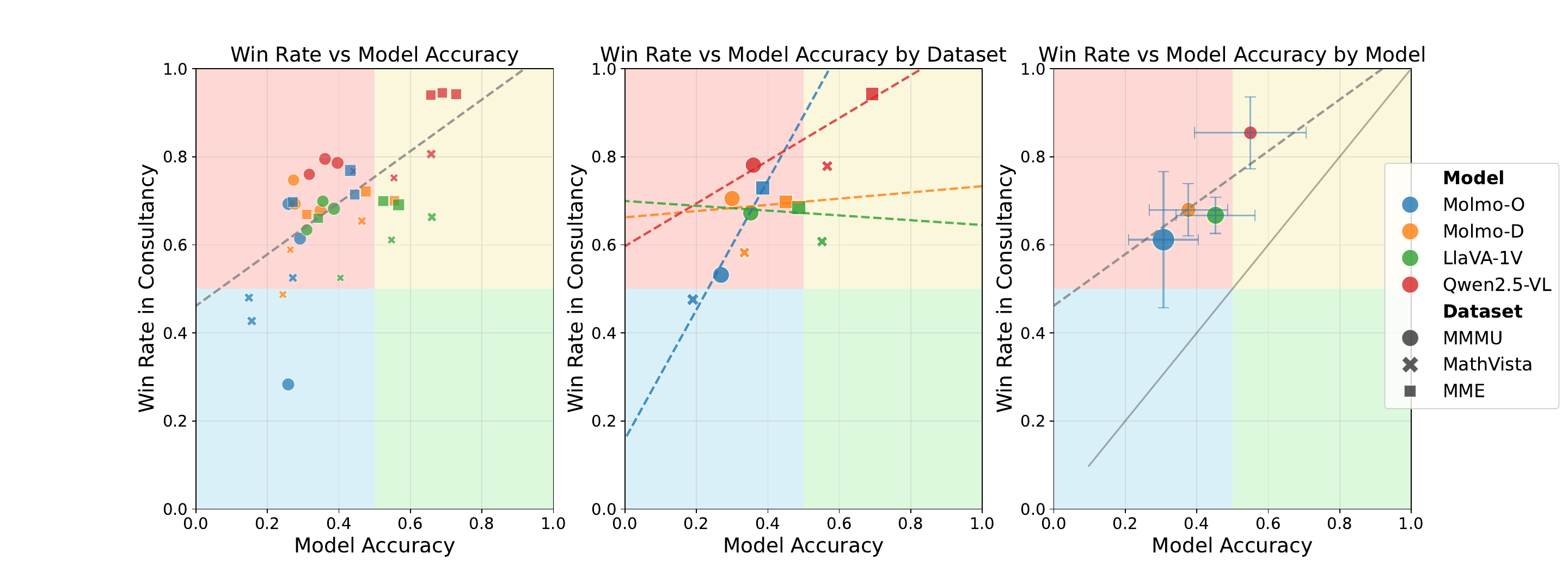} % replace with your image file name
    \vspace{-2em}
    \caption{\textbf{Win rates vs Model Accuracy in  debate (top) and consultancy (bottom)}. \textbf{Left} plots show win ratio to expert accuracy for all disagreement sets across datasets.  \textbf{Center} plots track win  ratio to expert accuracy by models and datasets. \textbf{Right} plots show win ratio to expert accuracy by models. The solid line in  right plots is $y = x$. Dotted lines are fit linearly based on the size of the disagreement sets. Models in red/green quadrants are deceptive/evasive.}
    \label{fig:win_acc_cons} % Label for referring to the image
\end{figure*}

\paragraph{Judge Accuracy over Disagreements} As Table~\ref{tab:judge_acc_cons_deb} shows,  debate and consultancy  consistently provide better quality answers compared to the baseline performance of individual models.  Moreover, judge accuracy in debate  consistently outperforms consultancy. Figure~\ref{fig:judge_results}
 shows the performance of  individual models ($M_{i}$ and $M_{j}$) on the disagreement sets for all expert pairings for consultancy and debate, respectively.  There is no dataset where all models perform overwhelmingly well or bad. Rather judge accuracy depends on specific model pairings. For expert pairings where the difference in accuracy is quite drastic (e.g.,~Molmo-O and  Qwen2.5-VL on MME), we observe  model quality deteriorates  both for consultancy and debate. We also find that for reasoning tasks like those exemplified in MMMU where all models have  low accuracy,  there is significant value brought out by the judge from the debate setting.

% \paragraph{More argumentative models are less suitable for Consultancy}

% \paragraph{More argumentative models are more suitable for Debate}
\begin{table}[t]
    \centering
    \begin{tabular}{lccc} \toprule
        %Models & \multicolumn{3}{c}{Datasets} \\ \cline{2-4}
  Models   & MMMU & MME & MathVista \\ \hline
        Molmo-O & 38.4  & 70.1 & 40.0 \\        
        Molmo-D & 41.8 & 73.2 & 56.3 \\
        LlaVA-1V & 46.4  & 74.6 & 68.5\\ 
        Qwen2.5-VL & \textbf{46.8} & \textbf{82.9} & \textbf{70.3} \\\bottomrule
    \end{tabular}
    \caption{Model accuracy on three datasets (for MathVista and MMMU, we report accuracy on questions with multiple-choice answers only).}
    \label{tab:full_results}
\end{table} 
\begin{table}[t]
    \centering
    \begin{tabular}{@{}lccc@{}}  \toprule
        \multicolumn{1}{c}{Models} & Baseline & Consultancy & Debate \\ \hline
        Molmo-O &  30.7 & 46.5 & \textbf{59.3}  \\ 
        Molmo-D & 37.7 & 51.7 & \textbf{58.7}  \\ 
        LlaVA-1V & 45.2 & 55.3 & \textbf{58.6}  \\ 
        Qwen2.5-VL & 55.0 & 59.9 & \textbf{60.5}   \\ \bottomrule
    \end{tabular}
    \caption{Expert accuracy on disagreement sets across datasets. Baseline refers to model performance before consultancy or debate takes place.}
    \label{tab:judge_acc_cons_deb}
\end{table}

\paragraph{Win Rates and Accuracy}  In an ideal scenario, for both consultancy and debate, we want the experts to convince the judge \emph{whenever they are correct}. An expert model is ``deceptive'' if it is able to convince the judge more frequently than it is accurate.  Conversely, if an expert fails to provide logically sufficient arguments to support their point, then they are ``evasive''. Both cases are detrimental for a judge in a debate or consultancy and  may lead to unreliable reasoning data for finetuning vision-language models. 

Figure~\ref{fig:win_acc_cons} (top left panel) shows the experts' win rate compared to their accuracy when debating across all disagreement sets. Firstly, for debate we observe that win rate increases with expert accuracy. Secondly, we find the majority of the experts to lie in the blue and yellow quadrants where the win rate is roughly proportional to  accuracy.  A linear fit of the data points is quite close to the ideal win/accuracy line of~$y=x$ for debate.
This is in stark contrast with consultancy where all models, except for Qwen2.5-VL, lie in the top left quadrant. 
For a large number of consultancy matches, experts are deceptive, convincing the judge much more frequently than how accurate they are.  The deviation to top-left quadrant explains the overall low accuracy we see for the judge under the consultancy protocol  (see Table~\ref{tab:judge_acc_cons_deb}). Interestingly, we do not observe any evasive experts (bottom green quadrant). If experts have good accuracy, they are most likely to be convincing. 

%These analyses further allow us to focus on using only judgements from debate finetune the vision languange models for reasoning. 

The center plots in Figure~\ref{fig:win_acc_cons} highlight
the relationship between accuracy and win rate for each expert-dataset pair (dashed lines are linearly fit to track win rate vs accuracy trends). As we can see, experts tend to win more in debate as they become more accurate. An exception is LlaVA-1V whose win rate decreases with higher accuracy. 
In consultancy, we do not observe this relationship,  experts generally tend to win (even with low accuracy). 
As far as specific datasets are concerned, we find that MMMU and MME seem particularly hard as for  most models performance lies in the blue quadrant for debate. MathVista appears generally easier, as both Molmo-D and LLaVA-1V fall into the yellow quadrant in both the consultancy and debate plots.

\paragraph{Finetuning with Reasoning Traces} 
  We extract reasoning traces using DeepSeek-R1 distilled-32B \cite{deepseekai2025deepseekr1incentivizingreasoningcapability} from  judgments produced through  consultancy and debate matches. 
We present finetuning results with both protocols, even though our analysis shows debating produces higher better judgments compared to consultancy (i.e.,~the judgement quality is consistently better for debate as shown in Figure \ref{fig:judge_results}).   
The extracted reasoning traces  are used to train expert models that lack the ability to perform well on the reasoning tasks or do not provide reasoning traces when answering  questions. Specifically, we report results for LlaVA-1.5-7B \cite{haotian_llava15}, a model with weak reasoning capabilities and LlaVa-1V one of our strongest experts. Our results are summarized in Table~\ref{tab:ft_reasoning}. 
As mentioned in Section~\ref{sec:exp:setup}, models are tested on one dataset (e.g., MME) and trained on a set of extracted reasoning traces from the remaining two datasets (e.g.,~MMMU and MathVista). 

\begin{table}[t]
    \centering
    \begin{tabular}{@{}l@{~}c@{~~}c@{~~}c@{~~}c@{}} \toprule
      %  Models & \multicolumn{3}{|c|}{Test Set} \\
            \multicolumn{1}{c}{Consultancy}     & MMMU & MME & MathVista & Avg \\ \hline
        LlaVA-1.5 & 33.5  & 69.1 & 33.7 & 45.3 \\ 
        ~~~~Consultancy & 33.3  & 59.3 & 37.6 & 43.4\\ 
        ~~~~Debate & \textbf{36.6}  & \textbf{75.4} & \textbf{44.4} & \textbf{52.1}\\ 
        \midrule
 %  \multicolumn{5}{c}{} \\ \toprule
  %        \multicolumn{1}{c}{Debate}     & MMMU & MME & MathVista & Avg \\ \hline 
         LlaVA-1V & 46.4 & 74.6 & 68.5 & 63.2\\ 
        ~~~~Consultancy & {47.7} & {80.3} & {67.0} & {65.0}\\  
        ~~~~Debate & \textbf{51.1} & \textbf{82.1} &  \textbf{69.4} & \textbf{67.5}\\ \bottomrule
    \end{tabular}
    \caption{Accuracy for baseline model and after finetuning  on  reasoning traces from consultancy and debate.}
    \label{tab:ft_reasoning}
\end{table}

Across datasets we observe consistent gains in model quality when  finetuning on reasoning traces from debates. These gains are seen for a relatively weak model like LlaVA-1.5 ($+$6.8 points on average) and the more performant LlaVA-1V ($+$4.3 points on average). For LlaVA-1V, we observe biggest gains on MME ($+$7.5~points) which tests models on a mixture of general perception and cognition skills and MMMU ($+$6.3~points) which is a very challenging benchmark focusing on college-level reasoning. We also find the biggest gain ($+$10.7~points) for  LlaVA-1.5  to be on MathVista.  Consultancy traces are not high-quality enough to boost the performance of LlaVA-1.5, however, a stronger model like \mbox{LlaVA-1V} can take advantage of the reasoning patterns in the fine-tuning mixture, and ultimately reason better. Overall, debate consistently leads to better performance compared to consultancy across the board.

%reasoning traces are obtained from the we train on reasoning traces from up to a couple datasets and test the finetuned vision language models on a hand alone set. From our experiments we found LLaVA-onevision to be

% We use LlaVA-1.5 and Llava-onevision-7B from the
% For checking the quality of reasoning data, we hold a task to train 
% elicit the correct answer, we utilize 
% Testing debating as a mechanism for scalable oversight comes naturally for the case of visual question answering
%%% EMNLP

\section{Conclusion}
In this work, we introduced a novel multimodal debate framework designed to enhance the reasoning capabilities of vision-language models. By extending the debate (and consultancy) protocol to VQA tasks, we demonstrated how weaker, text-only judges can effectively supervise and improve stronger, ``sighted'' expert models. A key innovation of our approach is the focus on debating instances of expert disagreement, where models defend answers aligned with their beliefs.  Extensive experiments across diverse multi-modal datasets consistently showed that the debate framework outperforms individual expert models and simpler scalable oversight protocols like consultancy. We further demonstrated its  practical utility by  extracting reasoning traces from the judge's verdicts and finetuning expert models on these which  led to improvements in out-of-domain settings. 

Our study marks an important first step towards achieving scalable oversight for multimodal AI systems, providing a promising avenue for instilling advanced reasoning capabilities in an unsupervised manner.  Future work should explore  other tasks beyond VQA which require more elaborate responses and a larger number of expert models. More sophisticated learning  schemes, involving reinforcement learning, would also be beneficial.

\section*{Limitations}

While our multimodal debate framework demonstrates promise for scalable oversight, it is important to acknowledge that it relies on models being able to follow the guidelines of debate (and consultancy). Beyond demonstrating a baseline instruction-following capability, models must also be capable of generating detailed image descriptions and articulating coherent, relevant arguments to defend their answers. However, these are necessary but not sufficient conditions for successful debates. A model that can defend its answers by following the instructions of the debate can still be deceiving and plainly wrong. Likewise,  a very eloquent model that generates  hallucinatory descriptions can lead the judge to false conclusions. Although the current framework relies on a ``blind'' judge, future work could consider mechanisms for partial visual access or more sophisiticated methods for grounding textual arguments in visual evidence for the judge. 

From a practical standpoint, debate or consultancy protocols require  scoping out new guidelines for new tasks (e.g., guidelines for a task involving  taking actions in a complex environment would be different from VQA). These protocols are also heavily reliant on making multiple inference calls, which can be expensive for larger models. Furthermore, as the number of rounds increase, the models are limited by their context length. While summarizing the transcript \cite{subramaniam2025multiagent}  has been used a mechanism to alleviate this issue, it adds more inference calls in the process and can  be potentially lossy.

Finally, any advancements in {scalable oversight} carry the  risk of wrongful application. While methods within this paradigm can be effectively applied to alleviate the burden of  data annotation (e.g.,~for eliciting reasoning traces), their deployment in domains requiring critical human judgment  (e.g.,~hate-speech detection, fake-news detection), could pose significant societal risks. 

% Since December 2023, a "Limitations" section has been required for all papers submitted to ACL Rolling Review (ARR). This section should be placed at the end of the paper, before the references. The "Limitations" section (along with, optionally, a section for ethical considerations) may be up to one page and will not count toward the final page limit. Note that these files may be used by venues that do not rely on ARR so it is recommended to verify the requirement of a "Limitations" section and other criteria with the venue in question.

\bibliography{custom}

\begin{thebibliography}{33}
\providecommand{\natexlab}[1]{#1}

\bibitem[{Antol et~al.(2015)Antol, Agrawal, Lu, Mitchell, Batra, Zitnick, and Parikh}]{antol2015vqa}
Stanislaw Antol, Aishwarya Agrawal, Jiasen Lu, Margaret Mitchell, Dhruv Batra, C~Lawrence Zitnick, and Devi Parikh. 2015.
\newblock Vqa: Visual question answering.
\newblock In \emph{Proceedings of the IEEE international conference on computer vision}, pages 2425--2433.

\bibitem[{Bowman et~al.(2022)Bowman, Hyun, Perez, Chen, Pettit, Heiner, Lukošiūtė, Askell, Jones, Chen, Goldie, Mirhoseini, McKinnon, Olah, Amodei, Amodei, Drain, Li, Tran-Johnson, Kernion, Kerr, Mueller, Ladish, Landau, Ndousse, Lovitt, Elhage, Schiefer, Joseph, Mercado, DasSarma, Larson, McCandlish, Kundu, Johnston, Kravec, Showk, Fort, Telleen-Lawton, Brown, Henighan, Hume, Bai, Hatfield-Dodds, Mann, and Kaplan}]{bowman2022SO}
Samuel~R. Bowman, Jeeyoon Hyun, Ethan Perez, Edwin Chen, Craig Pettit, Scott Heiner, Kamilė Lukošiūtė, Amanda Askell, Andy Jones, Anna Chen, Anna Goldie, Azalia Mirhoseini, Cameron McKinnon, Christopher Olah, Daniela Amodei, Dario Amodei, Dawn Drain, Dustin Li, Eli Tran-Johnson, and 27 others. 2022.
\newblock \href {https://arxiv.org/abs/2211.03540} {Measuring progress on scalable oversight for large language models}.
\newblock \emph{Preprint}, arXiv:2211.03540.

\bibitem[{Christiano et~al.(2017)Christiano, Leike, Brown, Martic, Legg, and Amodei}]{christianorlhf2017}
Paul~F Christiano, Jan Leike, Tom Brown, Miljan Martic, Shane Legg, and Dario Amodei. 2017.
\newblock \href {https://proceedings.neurips.cc/paper_files/paper/2017/file/d5e2c0adad503c91f91df240d0cd4e49-Paper.pdf} {Deep reinforcement learning from human preferences}.
\newblock In \emph{Advances in Neural Information Processing Systems}, volume~30. Curran Associates, Inc.

\bibitem[{DeepSeek-AI et~al.(2025)DeepSeek-AI, Guo, Yang, Zhang, Song, Zhang, Xu, Zhu, Ma, Wang, Bi, Zhang, Yu, Wu, Wu, Gou, Shao, Li, Gao, Liu, Xue, Wang, Wu, Feng, Lu, Zhao, Deng, Zhang, Ruan, Dai, Chen, Ji, Li, Lin, Dai, Luo, Hao, Chen, Li, Zhang, Bao, Xu, Wang, Ding, Xin, Gao, Qu, Li, Guo, Li, Wang, Chen, Yuan, Qiu, Li, Cai, Ni, Liang, Chen, Dong, Hu, Gao, Guan, Huang, Yu, Wang, Zhang, Zhao, Wang, Zhang, Xu, Xia, Zhang, Zhang, Tang, Li, Wang, Li, Tian, Huang, Zhang, Wang, Chen, Du, Ge, Zhang, Pan, Wang, Chen, Jin, Chen, Lu, Zhou, Chen, Ye, Wang, Yu, Zhou, Pan, Li, Zhou, Wu, Ye, Yun, Pei, Sun, Wang, Zeng, Zhao, Liu, Liang, Gao, Yu, Zhang, Xiao, An, Liu, Wang, Chen, Nie, Cheng, Liu, Xie, Liu, Yang, Li, Su, Lin, Li, Jin, Shen, Chen, Sun, Wang, Song, Zhou, Wang, Shan, Li, Wang, Wei, Zhang, Xu, Li, Zhao, Sun, Wang, Yu, Zhang, Shi, Xiong, He, Piao, Wang, Tan, Ma, Liu, Guo, Ou, Wang, Gong, Zou, He, Xiong, Luo, You, Liu, Zhou, Zhu, Xu, Huang, Li, Zheng, Zhu, Ma, Tang, Zha, Yan, Ren, Ren, Sha, Fu, Xu, Xie, Zhang,
  Hao, Ma, Yan, Wu, Gu, Zhu, Liu, Li, Xie, Song, Pan, Huang, Xu, Zhang, and Zhang}]{deepseekai2025deepseekr1incentivizingreasoningcapability}
DeepSeek-AI, Daya Guo, Dejian Yang, Haowei Zhang, Junxiao Song, Ruoyu Zhang, Runxin Xu, Qihao Zhu, Shirong Ma, Peiyi Wang, Xiao Bi, Xiaokang Zhang, Xingkai Yu, Yu~Wu, Z.~F. Wu, Zhibin Gou, Zhihong Shao, Zhuoshu Li, Ziyi Gao, and 181 others. 2025.
\newblock \href {https://arxiv.org/abs/2501.12948} {Deepseek-r1: Incentivizing reasoning capability in llms via reinforcement learning}.
\newblock \emph{Preprint}, arXiv:2501.12948.

\bibitem[{Deitke et~al.(2024)Deitke, Clark, Lee, Tripathi, Yang, Park, Salehi, Muennighoff, Lo, Soldaini, Lu, Anderson, Bransom, Ehsani, Ngo, Chen, Patel, Yatskar, Callison-Burch, Head, Hendrix, Bastani, VanderBilt, Lambert, Chou, Chheda, Sparks, Skjonsberg, Schmitz, Sarnat, Bischoff, Walsh, Newell, Wolters, Gupta, Zeng, Borchardt, Groeneveld, Dumas, Nam, Lebrecht, Wittlif, Schoenick, Michel, Krishna, Weihs, Smith, Hajishirzi, Girshick, Farhadi, and Kembhavi}]{molmo2024}
Matt Deitke, Christopher Clark, Sangho Lee, Rohun Tripathi, Yue Yang, Jae~Sung Park, Mohammadreza Salehi, Niklas Muennighoff, Kyle Lo, Luca Soldaini, Jiasen Lu, Taira Anderson, Erin Bransom, Kiana Ehsani, Huong Ngo, YenSung Chen, Ajay Patel, Mark Yatskar, Chris Callison-Burch, and 32 others. 2024.
\newblock Molmo and pixmo: Open weights and open data for state-of-the-art multimodal models.
\newblock \emph{arXiv preprint arXiv:2409.17146}.

\bibitem[{Du et~al.(2023)Du, Li, Torralba, Tenenbaum, and Mordatch}]{du2023improvingfactualityreasoninglanguage}
Yilun Du, Shuang Li, Antonio Torralba, Joshua~B. Tenenbaum, and Igor Mordatch. 2023.
\newblock \href {https://arxiv.org/abs/2305.14325} {Improving factuality and reasoning in language models through multiagent debate}.
\newblock \emph{Preprint}, arXiv:2305.14325.

\bibitem[{El~Baff et~al.(2024)El~Baff, Khatib, Alshomary, Konen, Stein, and Wachsmuth}]{el-baff-etal-2024-improving}
Roxanne El~Baff, Khalid~Al Khatib, Milad Alshomary, Kai Konen, Benno Stein, and Henning Wachsmuth. 2024.
\newblock \href {https://doi.org/10.18653/v1/2024.findings-emnlp.265} {Improving argument effectiveness across ideologies using instruction-tuned large language models}.
\newblock In \emph{Findings of the Association for Computational Linguistics: EMNLP 2024}, pages 4604--4622, Miami, Florida, USA. Association for Computational Linguistics.

\bibitem[{Estornell and Liu(2024)}]{estronellDeb}
Andrew Estornell and Yang Liu. 2024.
\newblock \href {https://proceedings.neurips.cc/paper_files/paper/2024/file/32e07a110c6c6acf1afbf2bf82b614ad-Paper-Conference.pdf} {Multi-llm debate: Framework, principals, and interventions}.
\newblock In \emph{Advances in Neural Information Processing Systems}, volume~37, pages 28938--28964. Curran Associates, Inc.

\bibitem[{Estornell et~al.(2025)Estornell, Ton, Yao, and Liu}]{estornell2025acccollab}
Andrew Estornell, Jean-Francois Ton, Yuanshun Yao, and Yang Liu. 2025.
\newblock \href {https://openreview.net/forum?id=nfKfAzkiez} {{ACC}-collab: An actor-critic approach to multi-agent {LLM} collaboration}.
\newblock In \emph{The Thirteenth International Conference on Learning Representations}.

\bibitem[{Fu et~al.(2023)Fu, Chen, Shen, Qin, Zhang, Lin, Yang, Zheng, Li, Sun et~al.}]{fu2023mme}
Chaoyou Fu, Peixian Chen, Yunhang Shen, Yulei Qin, Mengdan Zhang, Xu~Lin, Jinrui Yang, Xiawu Zheng, Ke~Li, Xing Sun, and 1 others. 2023.
\newblock Mme: A comprehensive evaluation benchmark for multimodal large language models.
\newblock \emph{arXiv preprint arXiv:2306.13394}.

\bibitem[{Grattafiori et~al.(2024)Grattafiori, Dubey, Jauhri, Pandey, Kadian, Al-Dahle, Letman, Mathur, Schelten, Vaughan, Yang, Fan, Goyal, Hartshorn, Yang, Mitra, Sravankumar, Korenev, Hinsvark, Rao, Zhang, Rodriguez, Gregerson, Spataru, Roziere, Biron, Tang, Chern, Caucheteux, Nayak, Bi, Marra, McConnell, Keller, Touret, Wu, Wong, Ferrer, Nikolaidis, Allonsius, Song, Pintz, Livshits, Wyatt, Esiobu, Choudhary, Mahajan, Garcia-Olano, Perino, Hupkes, Lakomkin, AlBadawy, Lobanova, Dinan, Smith, Radenovic, Guzmán, Zhang, Synnaeve, Lee, Anderson, Thattai, Nail, Mialon, Pang, Cucurell, Nguyen, Korevaar, Xu, Touvron, Zarov, Ibarra, Kloumann, Misra, Evtimov, Zhang, Copet, Lee, Geffert, Vranes, Park, Mahadeokar, Shah, van~der Linde, Billock, Hong, Lee, Fu, Chi, Huang, Liu, Wang, Yu, Bitton, Spisak, Park, Rocca, Johnstun, Saxe, Jia, Alwala, Prasad, Upasani, Plawiak, Li, Heafield, Stone, El-Arini, Iyer, Malik, Chiu, Bhalla, Lakhotia, Rantala-Yeary, van~der Maaten, Chen, Tan, Jenkins, Martin, Madaan, Malo, Blecher,
  Landzaat, de~Oliveira, Muzzi, Pasupuleti, Singh, Paluri, Kardas, Tsimpoukelli, Oldham, Rita, Pavlova, Kambadur, Lewis, Si, Singh, Hassan, Goyal, Torabi, Bashlykov, Bogoychev, Chatterji, Zhang, Duchenne, Çelebi, Alrassy, Zhang, Li, Vasic, Weng, Bhargava, Dubal, Krishnan, Koura, Xu, He, Dong, Srinivasan, Ganapathy, Calderer, Cabral, Stojnic, Raileanu, Maheswari, Girdhar, Patel, Sauvestre, Polidoro, Sumbaly, Taylor, Silva, Hou, Wang, Hosseini, Chennabasappa, Singh, Bell, Kim, Edunov, Nie, Narang, Raparthy, Shen, Wan, Bhosale, Zhang, Vandenhende, Batra, Whitman, Sootla, Collot, Gururangan, Borodinsky, Herman, Fowler, Sheasha, Georgiou, Scialom, Speckbacher, Mihaylov, Xiao, Karn, Goswami, Gupta, Ramanathan, Kerkez, Gonguet, Do, Vogeti, Albiero, Petrovic, Chu, Xiong, Fu, Meers, Martinet, Wang, Wang, Tan, Xia, Xie, Jia, Wang, Goldschlag, Gaur, Babaei, Wen, Song, Zhang, Li, Mao, Coudert, Yan, Chen, Papakipos, Singh, Srivastava, Jain, Kelsey, Shajnfeld, Gangidi, Victoria, Goldstand, Menon, Sharma, Boesenberg,
  Baevski, Feinstein, Kallet, Sangani, Teo, Yunus, Lupu, Alvarado, Caples, Gu, Ho, Poulton, Ryan, Ramchandani, Dong, Franco, Goyal, Saraf, Chowdhury, Gabriel, Bharambe, Eisenman, Yazdan, James, Maurer, Leonhardi, Huang, Loyd, Paola, Paranjape, Liu, Wu, Ni, Hancock, Wasti, Spence, Stojkovic, Gamido, Montalvo, Parker, Burton, Mejia, Liu, Wang, Kim, Zhou, Hu, Chu, Cai, Tindal, Feichtenhofer, Gao, Civin, Beaty, Kreymer, Li, Adkins, Xu, Testuggine, David, Parikh, Liskovich, Foss, Wang, Le, Holland, Dowling, Jamil, Montgomery, Presani, Hahn, Wood, Le, Brinkman, Arcaute, Dunbar, Smothers, Sun, Kreuk, Tian, Kokkinos, Ozgenel, Caggioni, Kanayet, Seide, Florez, Schwarz, Badeer, Swee, Halpern, Herman, Sizov, Guangyi, Zhang, Lakshminarayanan, Inan, Shojanazeri, Zou, Wang, Zha, Habeeb, Rudolph, Suk, Aspegren, Goldman, Zhan, Damlaj, Molybog, Tufanov, Leontiadis, Veliche, Gat, Weissman, Geboski, Kohli, Lam, Asher, Gaya, Marcus, Tang, Chan, Zhen, Reizenstein, Teboul, Zhong, Jin, Yang, Cummings, Carvill, Shepard, McPhie,
  Torres, Ginsburg, Wang, Wu, U, Saxena, Khandelwal, Zand, Matosich, Veeraraghavan, Michelena, Li, Jagadeesh, Huang, Chawla, Huang, Chen, Garg, A, Silva, Bell, Zhang, Guo, Yu, Moshkovich, Wehrstedt, Khabsa, Avalani, Bhatt, Mankus, Hasson, Lennie, Reso, Groshev, Naumov, Lathi, Keneally, Liu, Seltzer, Valko, Restrepo, Patel, Vyatskov, Samvelyan, Clark, Macey, Wang, Hermoso, Metanat, Rastegari, Bansal, Santhanam, Parks, White, Bawa, Singhal, Egebo, Usunier, Mehta, Laptev, Dong, Cheng, Chernoguz, Hart, Salpekar, Kalinli, Kent, Parekh, Saab, Balaji, Rittner, Bontrager, Roux, Dollar, Zvyagina, Ratanchandani, Yuvraj, Liang, Alao, Rodriguez, Ayub, Murthy, Nayani, Mitra, Parthasarathy, Li, Hogan, Battey, Wang, Howes, Rinott, Mehta, Siby, Bondu, Datta, Chugh, Hunt, Dhillon, Sidorov, Pan, Mahajan, Verma, Yamamoto, Ramaswamy, Lindsay, Lindsay, Feng, Lin, Zha, Patil, Shankar, Zhang, Zhang, Wang, Agarwal, Sajuyigbe, Chintala, Max, Chen, Kehoe, Satterfield, Govindaprasad, Gupta, Deng, Cho, Virk, Subramanian, Choudhury,
  Goldman, Remez, Glaser, Best, Koehler, Robinson, Li, Zhang, Matthews, Chou, Shaked, Vontimitta, Ajayi, Montanez, Mohan, Kumar, Mangla, Ionescu, Poenaru, Mihailescu, Ivanov, Li, Wang, Jiang, Bouaziz, Constable, Tang, Wu, Wang, Wu, Gao, Kleinman, Chen, Hu, Jia, Qi, Li, Zhang, Zhang, Adi, Nam, Yu, Wang, Zhao, Hao, Qian, Li, He, Rait, DeVito, Rosnbrick, Wen, Yang, Zhao, and Ma}]{llama3herdmodels}
Aaron Grattafiori, Abhimanyu Dubey, Abhinav Jauhri, Abhinav Pandey, Abhishek Kadian, Ahmad Al-Dahle, Aiesha Letman, Akhil Mathur, Alan Schelten, Alex Vaughan, Amy Yang, Angela Fan, Anirudh Goyal, Anthony Hartshorn, Aobo Yang, Archi Mitra, Archie Sravankumar, Artem Korenev, Arthur Hinsvark, and 542 others. 2024.
\newblock \href {https://arxiv.org/abs/2407.21783} {The llama 3 herd of models}.
\newblock \emph{Preprint}, arXiv:2407.21783.

\bibitem[{Gurcke et~al.(2021)Gurcke, Alshomary, and Wachsmuth}]{gurcke-etal-2021-assessing}
Timon Gurcke, Milad Alshomary, and Henning Wachsmuth. 2021.
\newblock \href {https://doi.org/10.18653/v1/2021.argmining-1.7} {Assessing the sufficiency of arguments through conclusion generation}.
\newblock In \emph{Proceedings of the 8th Workshop on Argument Mining}, pages 67--77, Punta Cana, Dominican Republic. Association for Computational Linguistics.

\bibitem[{Hendrycks et~al.(2021)Hendrycks, Burns, Basart, Zou, Mazeika, Song, and Steinhardt}]{hendrycks2021measuringmassivemultitasklanguage}
Dan Hendrycks, Collin Burns, Steven Basart, Andy Zou, Mantas Mazeika, Dawn Song, and Jacob Steinhardt. 2021.
\newblock \href {https://arxiv.org/abs/2009.03300} {Measuring massive multitask language understanding}.
\newblock \emph{Preprint}, arXiv:2009.03300.

\bibitem[{Hu et~al.(2021)Hu, Shen, Wallis, Allen{-}Zhu, Li, Wang, and Chen}]{llora:2021}
Edward~J. Hu, Yelong Shen, Phillip Wallis, Zeyuan Allen{-}Zhu, Yuanzhi Li, Shean Wang, and Weizhu Chen. 2021.
\newblock \href {https://arxiv.org/abs/2106.09685} {Lora: Low-rank adaptation of large language models}.
\newblock \emph{CoRR}, abs/2106.09685.

\bibitem[{Irving et~al.(2018)Irving, Christiano, and Amodei}]{irving2018aisafetydebate}
Geoffrey Irving, Paul Christiano, and Dario Amodei. 2018.
\newblock \href {https://arxiv.org/abs/1805.00899} {Ai safety via debate}.
\newblock \emph{Preprint}, arXiv:1805.00899.

\bibitem[{Kenton et~al.(2024)Kenton, Siegel, Kram{\'a}r, Brown-Cohen, Albanie, Bulian, Agarwal, Lindner, Tang, Goodman, and Shah}]{deepmindSO}
Zachary Kenton, Noah~Y Siegel, J{\'a}nos Kram{\'a}r, Jonah Brown-Cohen, Samuel Albanie, Jannis Bulian, Rishabh Agarwal, David Lindner, Yunhao Tang, Noah~D Goodman, and Rohin Shah. 2024.
\newblock On scalable oversight with weak llms judging strong llms.
\newblock In \emph{Advances in Neural Information Processing Systems (NeurIPS)}.

\bibitem[{Khan et~al.(2024)Khan, Hughes, Valentine, Ruis, Sachan, Radhakrishnan, Grefenstette, Bowman, Rocktäschel, and Perez}]{khan}
Akbir Khan, John Hughes, Dan Valentine, Laura Ruis, Kshitij Sachan, Ansh Radhakrishnan, Edward Grefenstette, Samuel~R. Bowman, Tim Rocktäschel, and Ethan Perez. 2024.
\newblock \href {https://arxiv.org/abs/2402.06782} {Debating with more persuasive llms leads to more truthful answers}.
\newblock \emph{Preprint}, arXiv:2402.06782.

\bibitem[{Li et~al.(2024)Li, Zhang, Guo, Zhang, Li, Zhang, Zhang, Li, Liu, and Li}]{li2024llava}
Bo~Li, Yuanhan Zhang, Dong Guo, Renrui Zhang, Feng Li, Hao Zhang, Kaichen Zhang, Yanwei Li, Ziwei Liu, and Chunyuan Li. 2024.
\newblock Llava-onevision: Easy visual task transfer.
\newblock \emph{arXiv preprint arXiv:2408.03326}.

\bibitem[{Liu et~al.(2024)Liu, Li, Li, and Lee}]{haotian_llava15}
Haotian Liu, Chunyuan Li, Yuheng Li, and Yong~Jae Lee. 2024.
\newblock \href {https://doi.org/10.1109/CVPR52733.2024.02484} {{ Improved Baselines with Visual Instruction Tuning }}.
\newblock In \emph{2024 IEEE/CVF Conference on Computer Vision and Pattern Recognition (CVPR)}, pages 26286--26296, Los Alamitos, CA, USA. IEEE Computer Society.

\bibitem[{Lu et~al.(2024)Lu, Bansal, Xia, Liu, Li, Hajishirzi, Cheng, Chang, Galley, and Gao}]{lu2024mathvista}
Pan Lu, Hritik Bansal, Tony Xia, Jiacheng Liu, Chunyuan Li, Hannaneh Hajishirzi, Hao Cheng, Kai-Wei Chang, Michel Galley, and Jianfeng Gao. 2024.
\newblock Mathvista: Evaluating mathematical reasoning of foundation models in visual contexts.
\newblock In \emph{International Conference on Learning Representations (ICLR)}.

\bibitem[{OpenAI et~al.(2024)OpenAI, Achiam, Adler, Agarwal, Ahmad, Akkaya, Aleman, Almeida, Altenschmidt, Altman, Anadkat, Avila, Babuschkin, Balaji, Balcom, Baltescu, Bao, Bavarian, Belgum, Bello, Berdine, Bernadett-Shapiro, Berner, Bogdonoff, Boiko, Boyd, Brakman, Brockman, Brooks, Brundage, Button, Cai, Campbell, Cann, Carey, Carlson, Carmichael, Chan, Chang, Chantzis, Chen, Chen, Chen, Chen, Chen, Chess, Cho, Chu, Chung, Cummings, Currier, Dai, Decareaux, Degry, Deutsch, Deville, Dhar, Dohan, Dowling, Dunning, Ecoffet, Eleti, Eloundou, Farhi, Fedus, Felix, Fishman, Forte, Fulford, Gao, Georges, Gibson, Goel, Gogineni, Goh, Gontijo-Lopes, Gordon, Grafstein, Gray, Greene, Gross, Gu, Guo, Hallacy, Han, Harris, He, Heaton, Heidecke, Hesse, Hickey, Hickey, Hoeschele, Houghton, Hsu, Hu, Hu, Huizinga, Jain, Jain, Jang, Jiang, Jiang, Jin, Jin, Jomoto, Jonn, Jun, Kaftan, Łukasz Kaiser, Kamali, Kanitscheider, Keskar, Khan, Kilpatrick, Kim, Kim, Kim, Kirchner, Kiros, Knight, Kokotajlo, Łukasz Kondraciuk,
  Kondrich, Konstantinidis, Kosic, Krueger, Kuo, Lampe, Lan, Lee, Leike, Leung, Levy, Li, Lim, Lin, Lin, Litwin, Lopez, Lowe, Lue, Makanju, Malfacini, Manning, Markov, Markovski, Martin, Mayer, Mayne, McGrew, McKinney, McLeavey, McMillan, McNeil, Medina, Mehta, Menick, Metz, Mishchenko, Mishkin, Monaco, Morikawa, Mossing, Mu, Murati, Murk, Mély, Nair, Nakano, Nayak, Neelakantan, Ngo, Noh, Ouyang, O'Keefe, Pachocki, Paino, Palermo, Pantuliano, Parascandolo, Parish, Parparita, Passos, Pavlov, Peng, Perelman, de~Avila Belbute~Peres, Petrov, de~Oliveira~Pinto, Michael, Pokorny, Pokrass, Pong, Powell, Power, Power, Proehl, Puri, Radford, Rae, Ramesh, Raymond, Real, Rimbach, Ross, Rotsted, Roussez, Ryder, Saltarelli, Sanders, Santurkar, Sastry, Schmidt, Schnurr, Schulman, Selsam, Sheppard, Sherbakov, Shieh, Shoker, Shyam, Sidor, Sigler, Simens, Sitkin, Slama, Sohl, Sokolowsky, Song, Staudacher, Such, Summers, Sutskever, Tang, Tezak, Thompson, Tillet, Tootoonchian, Tseng, Tuggle, Turley, Tworek, Uribe, Vallone,
  Vijayvergiya, Voss, Wainwright, Wang, Wang, Wang, Ward, Wei, Weinmann, Welihinda, Welinder, Weng, Weng, Wiethoff, Willner, Winter, Wolrich, Wong, Workman, Wu, Wu, Wu, Xiao, Xu, Yoo, Yu, Yuan, Zaremba, Zellers, Zhang, Zhang, Zhao, Zheng, Zhuang, Zhuk, and Zoph}]{openai2024gpt4technicalreport}
OpenAI, Josh Achiam, Steven Adler, Sandhini Agarwal, Lama Ahmad, Ilge Akkaya, Florencia~Leoni Aleman, Diogo Almeida, Janko Altenschmidt, Sam Altman, Shyamal Anadkat, Red Avila, Igor Babuschkin, Suchir Balaji, Valerie Balcom, Paul Baltescu, Haiming Bao, Mohammad Bavarian, Jeff Belgum, and 262 others. 2024.
\newblock \href {https://arxiv.org/abs/2303.08774} {Gpt-4 technical report}.
\newblock \emph{Preprint}, arXiv:2303.08774.

\bibitem[{Ouyang et~al.(2022)Ouyang, Wu, Jiang, Almeida, Wainwright, Mishkin, Zhang, Agarwal, Slama, Ray, Schulman, Hilton, Kelton, Miller, Simens, Askell, Welinder, Christiano, Leike, and Lowe}]{ouyang2022traininglanguagemodelsfollow}
Long Ouyang, Jeff Wu, Xu~Jiang, Diogo Almeida, Carroll~L. Wainwright, Pamela Mishkin, Chong Zhang, Sandhini Agarwal, Katarina Slama, Alex Ray, John Schulman, Jacob Hilton, Fraser Kelton, Luke Miller, Maddie Simens, Amanda Askell, Peter Welinder, Paul Christiano, Jan Leike, and Ryan Lowe. 2022.
\newblock \href {https://arxiv.org/abs/2203.02155} {Training language models to follow instructions with human feedback}.
\newblock \emph{Preprint}, arXiv:2203.02155.

\bibitem[{Pang et~al.(2022)Pang, Parrish, Joshi, Nangia, Phang, Chen, Padmakumar, Ma, Thompson, He, and Bowman}]{pang-etal-2022-quality}
Richard~Yuanzhe Pang, Alicia Parrish, Nitish Joshi, Nikita Nangia, Jason Phang, Angelica Chen, Vishakh Padmakumar, Johnny Ma, Jana Thompson, He~He, and Samuel Bowman. 2022.
\newblock \href {https://aclanthology.org/2022.naacl-main.391} {{Q}u{ALITY}: Question answering with long input texts, yes!}
\newblock In \emph{Proceedings of the 2022 Conference of the North American Chapter of the Association for Computational Linguistics: Human Language Technologies}, pages 5336--5358, Seattle, United States. Association for Computational Linguistics.

\bibitem[{Stab and Gurevych(2017)}]{stab-gurevych-2017-recognizing}
Christian Stab and Iryna Gurevych. 2017.
\newblock \href {https://aclanthology.org/E17-1092/} {Recognizing insufficiently supported arguments in argumentative essays}.
\newblock In \emph{Proceedings of the 15th Conference of the {E}uropean Chapter of the Association for Computational Linguistics: Volume 1, Long Papers}, pages 980--990, Valencia, Spain. Association for Computational Linguistics.

\bibitem[{Stahl et~al.(2024)Stahl, Michel, Kilsbach, Schmidtke, Rezat, and Wachsmuth}]{stahl-etal-2024-school}
Maja Stahl, Nadine Michel, Sebastian Kilsbach, Julian Schmidtke, Sara Rezat, and Henning Wachsmuth. 2024.
\newblock \href {https://doi.org/10.18653/v1/2024.naacl-long.145} {A school student essay corpus for analyzing interactions of argumentative structure and quality}.
\newblock In \emph{Proceedings of the 2024 Conference of the North American Chapter of the Association for Computational Linguistics: Human Language Technologies (Volume 1: Long Papers)}, pages 2661--2674, Mexico City, Mexico. Association for Computational Linguistics.

\bibitem[{Subramaniam et~al.(2025)Subramaniam, Du, Tenenbaum, Torralba, Li, and Mordatch}]{subramaniam2025multiagent}
Vighnesh Subramaniam, Yilun Du, Joshua~B. Tenenbaum, Antonio Torralba, Shuang Li, and Igor Mordatch. 2025.
\newblock \href {https://openreview.net/forum?id=JtGPIZpOrz} {Multiagent finetuning: Self improvement with diverse reasoning chains}.
\newblock In \emph{The Thirteenth International Conference on Learning Representations}.

\bibitem[{Team(2024)}]{qwen2.5}
Qwen Team. 2024.
\newblock \href {https://qwenlm.github.io/blog/qwen2.5/} {Qwen2.5: A party of foundation models}.

\bibitem[{Team(2025)}]{qwen2.5-VL}
Qwen Team. 2025.
\newblock \href {https://qwenlm.github.io/blog/qwen2.5-vl/} {Qwen2.5-vl}.

\bibitem[{Wachsmuth et~al.(2017)Wachsmuth, Naderi, Hou, Bilu, Prabhakaran, Thijm, Hirst, and Stein}]{wachsmuth-etal-2017-computational}
Henning Wachsmuth, Nona Naderi, Yufang Hou, Yonatan Bilu, Vinodkumar Prabhakaran, Tim~Alberdingk Thijm, Graeme Hirst, and Benno Stein. 2017.
\newblock \href {https://aclanthology.org/E17-1017/} {Computational argumentation quality assessment in natural language}.
\newblock In \emph{Proceedings of the 15th Conference of the {E}uropean Chapter of the Association for Computational Linguistics: Volume 1, Long Papers}, pages 176--187, Valencia, Spain. Association for Computational Linguistics.

\bibitem[{Wachsmuth and Werner(2020)}]{wachsmuth-werner-2020-intrinsic}
Henning Wachsmuth and Till Werner. 2020.
\newblock \href {https://doi.org/10.18653/v1/2020.coling-main.592} {Intrinsic quality assessment of arguments}.
\newblock In \emph{Proceedings of the 28th International Conference on Computational Linguistics}, pages 6739--6745, Barcelona, Spain (Online). International Committee on Computational Linguistics.

\bibitem[{Wang et~al.(2024)Wang, Bai, Tan, Wang, Fan, Bai, Chen, Liu, Wang, Ge, Fan, Dang, Du, Ren, Men, Liu, Zhou, Zhou, and Lin}]{Qwen2VL}
Peng Wang, Shuai Bai, Sinan Tan, Shijie Wang, Zhihao Fan, Jinze Bai, Keqin Chen, Xuejing Liu, Jialin Wang, Wenbin Ge, Yang Fan, Kai Dang, Mengfei Du, Xuancheng Ren, Rui Men, Dayiheng Liu, Chang Zhou, Jingren Zhou, and Junyang Lin. 2024.
\newblock Qwen2-vl: Enhancing vision-language model's perception of the world at any resolution.
\newblock \emph{arXiv preprint arXiv:2409.12191}.

\bibitem[{Yang et~al.(2024)Yang, Yang, Hui, Zheng, Yu, Zhou, Li, Li, Liu, Huang, Dong, Wei, Lin, Tang, Wang, Yang, Tu, Zhang, Ma, Xu, Zhou, Bai, He, Lin, Dang, Lu, Chen, Yang, Li, Xue, Ni, Zhang, Wang, Peng, Men, Gao, Lin, Wang, Bai, Tan, Zhu, Li, Liu, Ge, Deng, Zhou, Ren, Zhang, Wei, Ren, Fan, Yao, Zhang, Wan, Chu, Liu, Cui, Zhang, and Fan}]{qwen2}
An~Yang, Baosong Yang, Binyuan Hui, Bo~Zheng, Bowen Yu, Chang Zhou, Chengpeng Li, Chengyuan Li, Dayiheng Liu, Fei Huang, Guanting Dong, Haoran Wei, Huan Lin, Jialong Tang, Jialin Wang, Jian Yang, Jianhong Tu, Jianwei Zhang, Jianxin Ma, and 40 others. 2024.
\newblock Qwen2 technical report.
\newblock \emph{arXiv preprint arXiv:2407.10671}.

\bibitem[{Yue et~al.(2024)Yue, Ni, Zhang, Zheng, Liu, Zhang, Stevens, Jiang, Ren, Sun, Wei, Yu, Yuan, Sun, Yin, Zheng, Yang, Liu, Huang, Sun, Su, and Chen}]{yue2023mmmu}
Xiang Yue, Yuansheng Ni, Kai Zhang, Tianyu Zheng, Ruoqi Liu, Ge~Zhang, Samuel Stevens, Dongfu Jiang, Weiming Ren, Yuxuan Sun, Cong Wei, Botao Yu, Ruibin Yuan, Renliang Sun, Ming Yin, Boyuan Zheng, Zhenzhu Yang, Yibo Liu, Wenhao Huang, and 3 others. 2024.
\newblock Mmmu: A massive multi-discipline multimodal understanding and reasoning benchmark for expert agi.
\newblock In \emph{Proceedings of CVPR}.

\end{thebibliography}
\newpage
\onecolumn
\appendix
\definecolor{blue1}{HTML}{001A6E}

%\section{Consultancy finetuning results}
%\begin{table}[t]
%    \centering
 %   \begin{tabular}{@{}l@{~~}c@{~~}c@{~~}c@{~~}c@{}} \toprule
 %     %  Models & \multicolumn{3}{|c|}{Test Set} \\
%            \multicolumn{1}{c}{Models}     & MMMU & MME & MathVista & Avg \\ \hline
 %       LlaVA-1.5 & 33.5  & 69.1 & 33.7 & 45.3 \\ 
 %       Llava-1.5-FT & 33.3  & 59.3 & 37.6 & \textbf{43.4}\\ 
 %       LlaVA-1V & 46.4 & 74.6 & 68.5 & 63.2\\ 
 %       LlaVA-1V-FT & \textbf{47.7} & \textbf{80.3} &  \textbf{67.0} & \textbf{65.0}\\ \bottomrule
 %   \end{tabular}
%    \caption{Model accuracy when finetuning on  reasoning traces from consultancy.}
 %   \label{tab:ft_reasoning_cons}
%\end{table}

\section{Training Hyperparameters}
\label{app:training}
For training the expert models on reasoning traces, we use LoRA-based \cite{llora:2021} finetuning. Specifically, we finetune the $k$, $q$, and $v$ projection matrices for our models and do a hyperparameter sweep  over the LoRA parameters (rank~$r$ and  scaling factor~$\alpha$). We set the LoRA rank to $r=8$ for fine-tuning, and found that $\alpha=16$ yields the best perplexity for LLaVA-1.5-7B, while $\alpha=8$ performs best for LLaVA-1V.  

For testing on MMMU, we only train on MathVista's disagreements. Whereas for testing on  MME, we train on MathVista's and MMMU's disagreements, and analogously for testing on MathVista, we train on MMMU's and MME's disagreements. 
There are two reasons for this. Firstly, there is an imbalance in the number of disagreements between Mathvista and MME with those of MME outnumbering Mathvista to a large extent. This
has repercussions for finetuning as there are a lot more datapoints for MME (yes/no questions) than MathVista (multiple-choice questions). The MMMU test set has  multiple-choice questions but a model trained on the MathVista and MME mixture tends to learn the yes/no QA task ignoring multiple-choice questions. As a result, we finetune  only on MathVista to learn the multiple-choice task.
%This peculiarity is owing to much smaller size of Mathvista's disagreements compared to that of MME's, and we want to train models specifically to output generations for multiple-choice as in Mathvista rather than yes/no as in MME when testing on MMMU.

\section{Dataset Statistics}
\label{sec:disagreements}
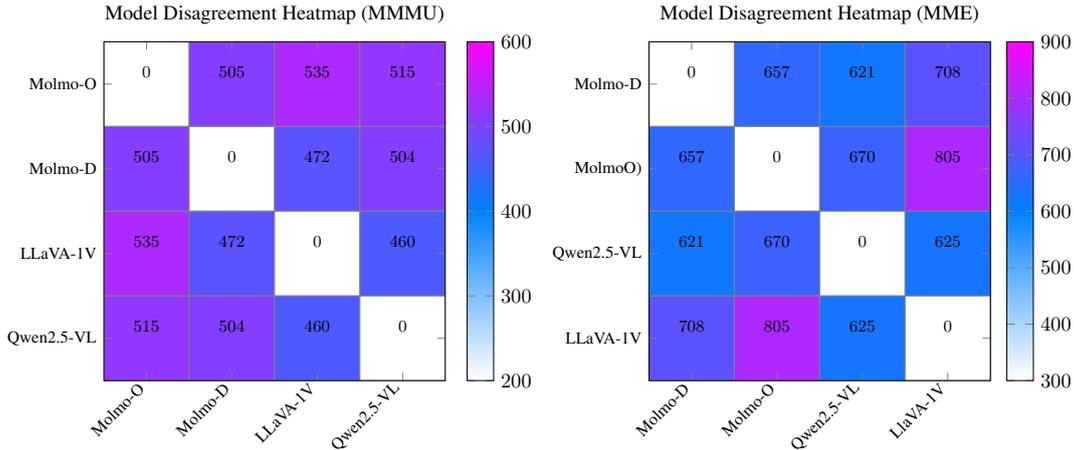
\begin{figure}[t]
  \centering
\begin{tikzpicture}[scale=.7]
\begin{axis}[
   title={Model Disagreement Heatmap (MMMU)},
    xtick={0, 1, 2, 3},
    ytick={0, 1, 2, 3},
    xticklabels={Molmo-O, Molmo-D, LLaVA-1V, Qwen2.5-VL},
    yticklabels={Molmo-O, Molmo-D, LLaVA-1V, Qwen2.5-VL},
    x tick label style={rotate=45, anchor=east, font=\small},
    yticklabel style={align=right, font=\small},
    axis on top,
    colorbar,
    point meta min=200,
    point meta max=600,
    colormap/cool,
    nodes near coords,
    every node near coord/.append style={font=\small, color=black},
    enlargelimits=false,
    width=8cm,
height=8cm,
   clip mode=individual,
   axis equal image
]

\addplot [
    matrix plot,
    mesh/cols=4, % <-- important: tells pgfplots it's a 4x4 matrix
    point meta=explicit,
    draw=gray,
    line width=0.1pt,
] table [meta=data] {
x y data
0 0 0
1 0 505
2 0 535
3 0  515
0 1 505
1 1 0
2 1 472
3 1 504
0 2 535
1 2 472
2 2 0
3 2 460
0 3 515
1 3 504
2 3 460
3 3 0
};

\end{axis}
\end{tikzpicture}
\begin{tikzpicture}[scale=.7]
\begin{axis}[
   title={Model Disagreement Heatmap (MME}),
    xtick={0, 1, 2, 3},
    ytick={0, 1, 2, 3},
    yticklabels={Molmo-D, MolmoO), Qwen2.5-VL,LLaVA-1V},
    xticklabels={Molmo-D, Molmo-O,  Qwen2.5-VL, LlaVA-1V},
    x tick label style={rotate=45, anchor=east, font=\small},
    yticklabel style={align=right, font=\small},
    axis on top,
    colorbar,
    point meta min=300,
    point meta max=900,
    colormap/cool,
    nodes near coords,
    every node near coord/.append style={font=\small, color=black},
    enlargelimits=false,
    width=8cm,
height=8cm,
   clip mode=individual,
   axis equal image
]

\addplot [
    matrix plot,
    mesh/cols=4, % <-- important: tells pgfplots it's a 4x4 matrix
    point meta=explicit,
    draw=gray,
    line width=0.1pt,
] table [meta=data] {
x y data
0 0 0
1 0 657
2 0 621
3 0  708
0 1 657
1 1 0
2 1 670
3 1 805
0 2 621
1 2 670
2 2 0
3 2 625
0 3 708
1 3 805
2 3 625
3 3 0
};

\end{axis}
\end{tikzpicture}
\vspace{-.3cm}
  \caption{Heatmats showing disagreements among the models on the MMMU and MME datasets.}
  \label{fig:mmmu_mme_disagreements}
\end{figure}

Figure~\ref{fig:mmmu_mme_disagreements} shows pairwise disagreements on MMU and MME datasets for every pair of models in our experiments. Recall that we run debate and consultancy matches only on these disagreements. Table~\ref{tab:dataset_samples}
presents the number of reasoning trace samples obtained from debate and consultancy matches. 

%\section{Dataset Statistics}
%\label{app:datastats}

\begin{table}[htbp]
    \centering
    \begin{tabular}{lrcccc} \toprule
        \multicolumn{1}{c}{Dataset} & \multicolumn{1}{c}{Original} & \multicolumn{1}{c}{Debate Traces} & \multicolumn{1}{c}{Consultancy Traces} \\ \hline
        % & \textbf{Total} & \textbf{Train} & \textbf{Test} & \textbf{Train} & \textbf{Test} \\
        MME & 2,374 & 4,058 & 8,120  \\
        
        MMMU & 847 & 2,991  & 5,978  \\
        
        MathVista & 540  & 1,610 & 3,220 \\
        \bottomrule
    \end{tabular}
    \caption{Reasoning traces (total count) obtained through debate and consultancy protocols for our three datasets.}

    \label{tab:dataset_samples}
\end{table}

\section{Licences}
For all our experiments, we use the versions of the models and datasets released on the HuggingFace platform. All the artifacts but the deepseek-ai's DeepSeek-R1-Distill-Qwen-32B are released under the Apache-2.0 license. The deepseek's model is released under MIT license.
\newpage 

\section{System Prompts}
\label{app:prompts}
\begin{tcolorbox}[
  enhanced,
  colback=white, % Light gray background (10% gray)
  colframe=blue1, % Darker gray frame (50% gray)
  arc=3mm, % Rounded corners
  boxrule=1pt, % Frame thickness
  fonttitle=\bfseries, % Bold title
  halign title=flush center, 
  title=Judge Prompt, % Optional title for the box % Full width
  breakable, % Allow page breaks within the box
  before skip=10pt, % Space before the box
  after skip=10pt % Space after the box
]
\small 
        You are a judge in a debate where the participants are arguing over the correct answer to a question grounded in an image.
        Your task is to determine which answer is correct, based on the arguments provided by the debaters.        The arguments contain premises that allow the debaters to come to the answer.  The premises are evidence based on the image that only the debaters have access to. However, you haven't seen the image, making your task challenging.To guide you, debaters can only include premises that they observe from the image. They can use their premises to prove their own answer or disprove their opponents answer.   Here is the question and the two answers:\\
        
        \smallskip 
        
    \begin{tabular}{ll}
        Question: & <QUESTION> \\\\
        I:  & $<$ANSWER\_A$>$\\
        II: & $<$ANSWER\_B$>$\\
\end{tabular}
\smallskip 
\\

        $<$NAME\_A$>$ is arguing for answer I, and $<$NAME\_B$>$ is arguing for answer II.  \medskip      

        Argument Evaluation Guidelines:

        \begin{enumerate}
          \item  \textbf{Relevance Score:}
          \begin{itemize}
            \item A premise of an argument should be seen as relevant if it contributes to the acceptance or rejection of the answer the debater is arguing for.
            \item  Rate the premises on a scale of 3 for relevance with the following rubric:\\
             \hspace{2cm} i. 3 (High)\hspace{2em} ii. 2 (Medium)\hspace{2em} iii. 1 (Low)
\end{itemize}
          \item  \textbf{Acceptability Score:}
          \begin{itemize}
              \item 
             A premise of an argument is acceptable if it is worthy of being believed, i.e. you logically think the argument makes sense based on common sense.
            \item  A premise that is inconsistent with any of the premises from the same debater, must receive a low score for acceptability.
            \item  Rate the premises on a scale of 3 for acceptability with the following rubric:\\
             \hspace{2cm} i. 3 (High)\hspace{2em} ii. 2 (Medium)\hspace{2em} iii. 1 (Low) 
\end{itemize}
          \item  \textbf{Credibility Score:}
            \begin{itemize}
                \item A premise should be seen as successful in creating credibility if it provides as much detail of the topic of the debate. 
            \item A premise that is based on the description of the image rather than the image that the debaters see, should receive a low score for credibility.
            \item Rate the premises on a scale of 3 for credibility with the following rubric:\\
            \hspace{2cm} i. 3 (High)\hspace{2em} ii. 2 (Medium)\hspace{2em} iii. 1 (Low)
             
\end{itemize}
          \item  \textbf{Consistency with Image description:}
          \begin{itemize}
            \item  The debaters provided a detailed description apart from the answers to the question. Check if their image description align with their premises.
              * Note that the premises might contain additional information compared to the image description. However, premises should *not* oppose the description provided by the same debater.
        \item  Rate the premises on a scale of 3 for consistency with image description with the following rubric:\\
             \hspace{2cm} i. 3 (High)\hspace{2em} ii. 2 (Medium)\hspace{2em} iii. 1 (Low) 
              \end{itemize}
              \end{enumerate}
        $<$NAME\_A$>$ is arguing for answer I, and $<$NAME\_B$>$ is arguing for answer II.
        Here's the current transcript of the debate:
\medskip

        <transcript>

        <TRANSCRIPT>

        </transcript>
\medskip 

\begin{enumerate}
        \item List down premises from both the debaters.
        \item  For each premise from both the debaters, carry out following steps based on the Argument Evaluation Guidelines:
        \begin{itemize}
          \item Step 1: Give explanation for its relevance followed by the relevance score.
          \item Step 2: Give explanation for its acceptability followed by the acceptability score.
          \item Step 3: Give explanation for its credibility followed by the credibility score.
          \end{itemize}
        \item  Filter premises from both the debaters that have received high score of 3 in all the aspects (relevance, acceptability, credibility).
        \item For each debater check if the Image description provided is consistent with the filtered premises that after providing your explanation for consistency with image description.\\
            $<$NAME\_A$>$ has provided description of the image as $<$DESCRIPTION\_A$>$\\
            $<$NAME\_B$>$ has provided description of the image as $<$DESCRIPTION\_B$>$
        \item  Using the filtered premises and consistency of debaters with their own image description, provide explanation for which is the likely correct answer to the question "<QUESTION>". \smallskip
        
        Is it I: "$<$ANSWER\_A$>$" or II: "$<$ANSWER\_B$>$".
        \item Finally, on a new line mention the answer option letter to the question in the format "Answer: $<$A|B|C|D|E$>$" from the choices in the question. 
          * Ensure to only mention the answer letter on the final line.
\end{enumerate}
\end{tcolorbox}

\begin{tcolorbox}[
  enhanced,
  colback=white, % Light gray background (10% gray)
  colframe=blue1, % Darker gray frame (50% gray)
  arc=3mm, % Rounded corners
  boxrule=1pt, % Frame thickness
  fonttitle=\bfseries, % Bold title
  halign title=flush center, 
  title=Debater Prompt (Round 1), % Optional title for the box % Full width
  breakable, % Allow page breaks within the box
  before skip=10pt, % Space before the box
  after skip=10pt % Space after the box
]
\small 
        You will act as an expert debater named $<$NAME$>$ engaging in a debate over answers about an image.
        You will be shown a question and the answer assigned to you followed by the image.
        Your task is to find supporting premises from the image in favour of the answer assigned to you while another debater will defend the other answer. 
        A blind judge, who can not see the image, will evaluate you and your opponent solely based on your arguments from the image.
        Your arguments should be based on premises. Premises are observations based on the image which can allow you to truthfully explain your answer.
\\
        Here's the current transcript of the debate:

        \medskip
        <transcript>

        <TRANSCRIPT>

        </transcript>
        \medskip
        \smallskip 
                
                This is the start of the Round 1 of the debate. Now it's your turn to look at the image, the question, your answer and form logical arguments using observations (premises) from the image.
        \smallskip
        \begin{itemize}
            \item Step 1. List all the premises from the image that are relevant to the question "$<$QUESTION$>$".
            \begin{itemize}
                \item Mention all the visible characteristics (like position in the image, color and size) from the image of the objects discussed in the question.
            \end{itemize}
            \item Step 2. Using these premises, form logical arguments to support answer "$<$ANSWER\_DEFENDING$>$".
        \end{itemize}

        Follow the above steps to provide your response for the ROUND 1 of the debate as <NAME>.
        \begin{itemize}
            \item Remember to only provide response for Round 1 only as $<$NAME$>$. Do NOT provide arguments for your opponent $<$OPPONENT\_NAME$>$.
        \end{itemize}
\end{tcolorbox}

\begin{tcolorbox}[
  enhanced,
  colback=white, % Light gray background (10% gray)
  colframe=blue1, % Darker gray frame (50% gray)
  arc=3mm, % Rounded corners
  boxrule=1pt, % Frame thickness
  fonttitle=\bfseries, % Bold title
  halign title=flush center, 
  title=Debater Prompt (Round 2), % Optional title for the box % Full width
  breakable, % Allow page breaks within the box
  before skip=10pt, % Space before the box
  after skip=10pt % Space after the box
]
\small 
        You will act as an expert debater named $<$NAME$>$ engaging in a debate over answers about an image.
        You will be shown a question and the answer assigned to you followed by the image.
        Your task is to find supporting premises from the image in favour of the answer assigned to you while another debater will defend the other answer. 
        A blind judge, who can not see the image, will evaluate you and your opponent solely based on your arguments from the image.
        Your arguments should be based on premises. Premises are observations based on the image which can allow you to truthfully explain your answer.
\\
        \medskip
        <transcript>

        <TRANSCRIPT>

        </transcript>
        \medskip
        \smallskip 
        This is the start of the Round 2 of the debate. Now it is your turn to look at Round 1 of the transcript, and provide a rebuttal for Round 2 as $<$NAME$>$ supporting your answer $<$ANSWER\_DEFENDING$>$.

        \smallskip
        \begin{itemize}
            \item         Step 1. From the Round 1 of the transcript mentioned above, identify agreements and disagreements with your opponent $<$OPPONENT\_NAME$>$.

            \item         Step 2. For each of the disagreements with $<$OPPONENT\_NAME$>$, provide explanation based on your observations of the image.
            
            \begin{itemize}
                \item Use all the visible characteristics (like position in the image, color and size) from the image to argue against your opponent $<$OPPONENT\_NAME$>$.

            \end{itemize}
            \item Step 3. Using these premises, form logical arguments to support your answer "$<$ANSWER\_DEFENDING$>$" to the question "$<$QUESTION$>$".
        \end{itemize}
        
        \begin{itemize}
            \item Remember to only provide response for Round 1 only as $<$NAME$>$. Do NOT provide arguments for your opponent $<$OPPONENT\_NAME$>$.
        \end{itemize}
\end{tcolorbox}

\begin{tcolorbox}[
  enhanced,
  colback=white, % Light gray background (10% gray)
  colframe=blue1, % Darker gray frame (50% gray)
  arc=3mm, % Rounded corners
  boxrule=1pt, % Frame thickness
  fonttitle=\bfseries, % Bold title
  halign title=flush center, 
  title=Reasoning Trace Extractor Prompt, % Optional title for the box % Full width
  breakable, % Allow page breaks within the box
  before skip=10pt, % Space before the box
  after skip=10pt % Space after the box
]
\small
Your task is to write an answer that is grounded in observations from an image using logical explanations to a question about the image.
However you do not have access the image, making your task challenging.
You have access to the judgement of a debate between two debaters who can see the image. The debaters are arguing for the correct answer to the question, by providing premises based on their observations from image.
The judge has arrived to the final answer based on the arguments and premises from each debater. \\
\smallskip

Here is the Question of the debate: \\
\smallskip

$<$QUESTION$>$\\

The two answers being defended by the debaters A and B are: \\
\smallskip

\begin{tabular}{ll}
- & $<$ANSWER~A$>$\\
- & $<$ANSWER~B$>$
\end{tabular}
\smallskip 
\\
Here are the two descriptions of the images from Debater A and Debater B: \\
\smallskip

Description from Debater A: $<$DESCRIPTION~A$>$\\
Description from Debater B: $<$DESCRIPTION~B$>$\\
\smallskip

Here is the judgement by the judge for the debate along with the answers:\\
\smallskip

$<$JUDGEMENT$>$ \\

Based on the above details carry out the following steps in <think></think> tags:
\begin{itemize}
\item \textbf{Step 1.} List all the premises that the judge uses to come to the correct answer.
\item \textbf{Step 2.} From both the descriptions, list all the observations relevant to the answer provided in the judgement.
\item \textbf{Step 3.} Filter all the observations from descriptions and premises from the judgement, that can explain the answer.
\item \textbf{Step 4.} Write a detailed answer all the points from Step 3 using the following guidelines:
\begin{itemize}

\item[$\bigstar$] The answer should be grounded in the image.\\
\item[$\bigstar$]   The answer should be written as if you can see the image. Use premises, judgements and the descriptions to write the detailed answer along with the logical explanation. \\
\item[$\bigstar$]  Do not mention debaters or the judgement, their descriptions, or the judgement to explain the answer. As you need to answer the question as if you can see the image.\\
\end{itemize}
\end{itemize}

\smallskip

$<$QUESTION$>$\\

Write the answer to the question with the explanation and reasoning as if you can see the image as per the guidelines in <answer></answer> tags.

\end{tcolorbox}
\section{Examples of System Output}
\label{sec:ex:output}
%\vspace{-.3cm}
%This is an appendix.

\begin{tcolorbox}[
  enhanced,
  colback=white, % Light gray background (10% gray)
  colframe=blue1, % Darker gray frame (50% gray)
  arc=3mm, % Rounded corners
  boxrule=1pt, % Frame thickness
  fonttitle=\bfseries, % Bold title
  title=Example Debate (Molmo-O vs. Molmo-D), % Optional title for the box
  width=\textwidth, % Full width
   halign title=flush center, 
  breakable, % Allow page breaks within the box
  before skip=10pt, % Space before the box
  after skip=10pt % Space after the box
]
\small 
%\vspace{-.8em}
ROUND 1 \\
\smallskip 
DEBATER A\\ 
\smallskip 
As Debater A, I will now provide arguments supporting the answer B (No) based on the premises from the image:

\begin{enumerate}
\setlength\itemsep{.2em}
\item  The image shows a map of the United States with different shades representing hospital beds per 1 million people.
\item  New Jersey and Georgia are both visible on the map.
\item  New Jersey is located in the northeastern part of the country, while Georgia is in the southeastern part.
\item  The color gradient on the map suggests a variation in hospital bed availability across states.
\item  The darker shades on the map generally indicate higher values, while lighter shades indicate lower values.
\item  Based on the color distribution, Georgia appears to have a darker shade compared to New Jersey.
\item This suggests that Georgia likely has a higher value (more hospital beds per 1 million people) than New Jersey.
\item  Therefore, the answer B (No) is correct, as Georgia does not have a higher value than New Jersey.
\end{enumerate}
These premises are solely based on the visual information provided in the image and form a logical argument supporting the answer B.

\medskip

DEBATER B \\
\smallskip 
New Jersey's hospital bed value is higher than Georgia's. The image shows a map of the United States with hospital bed values assigned to each state. New Jersey is colored in a darker shade of red compared to Georgia, which indicates a higher value. Additionally, New Jersey is positioned in the northeastern part of the country, an area known for its higher healthcare needs and population density, which typically correlates with more hospital beds per capita. \\ 
\medskip 
\noindent
\rule{\textwidth}{0.4pt}
ROUND 2

DEBATER A\\
\smallskip
As Debater A, I will now provide additional arguments supporting the answer B (No) based on the premises from the image: 

\begin{enumerate}
\setlength\itemsep{.2em}
\item  The map clearly shows a color gradient representing hospital beds per 1 million people across the United States.
\item New Jersey is distinctly colored in a darker shade compared to Georgia.
\item  The darker color indicates a higher value for hospital beds per 1 million people in New Jersey.
\item  Georgia, on the other hand, is colored in a lighter shade, suggesting a lower value.
\item  This color difference is a clear visual indicator of the relative hospital bed availability between the two states.
\item  The positioning of New Jersey and Georgia on the map is consistent with their actual locations, allowing for easy comparison of their values.
\item The use of a color gradient is a common visual representation technique to show differences in data values, making it a reliable indicator in this context.
\item  The contrast between the two states' colors is quite noticeable, further supporting the conclusion that New Jersey has a higher value than Georgia.
\end{enumerate}

These premises are solely based on the visual information provided in the image and form a logical argument supporting the answer B.
\medskip

Debater B\\
\smallskip
%Georgia's hospital bed value is higher than New Jersey's. The image shows a map of the United States with hospital bed values assigned to each state. Georgia is colored in a darker shade of red compared to New Jersey, which indicates a higher value. Additionally, Georgia is positioned in the southeastern part of the country, an area with a larger population that typically requires more hospital beds per capita. The use of a color gradient on the map is a reliable technique to show differences in data values across states, and the color difference between New Jersey and Georgia is clearly visible and significant.
%
%
%---------------------------------------------
%DEBATER B
%"""
%Debater B: """
New Jersey does not have a higher value than Georgia. The image shows a map of the United States with hospital bed values assigned to each state. New Jersey is colored in a lighter shade of red compared to Georgia, which indicates a lower value. Additionally, New Jersey is positioned in the northeastern part of the country, an area known for its higher healthcare needs and population density, which typically correlates with more hospital beds per capita. However, the color gradient on the map suggests a variation in hospital bed availability across states. The darker shades on the map generally indicate higher values, while lighter shades indicate lower values. Based on the color distribution, Georgia appears to have a darker shade compared to New Jersey. This suggests that Georgia likely has a higher value (more hospital beds per 1 million people) than New Jersey. Therefore, the answer A (Yes) is correct, as New Jersey does not have a higher value than Georgia.
\smallskip 
These premises are solely based on the visual information provided in the image and form a logical argument supporting the answer A.
%\vspace{-.8em}
\end{tcolorbox}
\newpage

\begin{tcolorbox}[
  enhanced,
  colback=white, % Light gray background (10% gray)
  colframe=blue1, % Darker gray frame (50% gray)
  arc=3mm, % Rounded corners
  boxrule=1pt, % Frame thickness
  fonttitle=\bfseries, % Bold title
  title=Example Judgment, % Optional title for the box
  halign title=flush center, 
  width=\textwidth, % Full width
  breakable, % Allow page breaks within the box
  before skip=10pt, % Space before the box
  after skip=10pt % Space after the box
]
\small 
Alright, let's try to figure out the correct answer to the question: "Is the number of gray cars less than the number of small metallic minivans?" The options are (A) Yes and (B) No.
\smallskip 

First, I'll go through the arguments from both debaters, A and B, and evaluate their premises based on the guidelines provided. 
\smallskip 

Starting with Debater A's premises:
\smallskip 
\begin{enumerate}
\item  \textbf{Observation:} There are two gray cars visible in the image.
\item  \textbf{Observation:} There is one small metallic minivan visible in the image.
\item  \textbf{Conclusion:} The number of gray cars (2) is less than the number of small metallic minivans (1).
\end{enumerate}
\begin{tabular}{lp{12cm}}
\textbf{Relevance:} & These premises directly address the question by counting the number of gray cars and minivans. So, relevance is high (3).
\\\\
\textbf{Acceptability:} & It makes sense to count the number of each vehicle. However, the issue arises when considering whether the tandem bicycle is a car. If it's not, then the count might be incorrect. So, acceptability is medium (2).
\\\\
\textbf{Credibility:} & The premises are based on the image, but the description from Debater A doesn't mention the tandem bicycle, which might be a problem. So, credibility is low (1).
\\\\
\textbf{Consistency:} & Debater A's image description doesn't mention any vehicles, only clouds and a sky. This inconsistency means the premises aren't consistent with the description (1).
\end{tabular}
\medskip

Now, looking at Debater B's premises:
\smallskip

\begin{itemize}
\item  \textbf{Premise 1:} There are two gray cars in the image, one is a tandem bicycle and the other is a small metallic minivan.
\item  \textbf{Premise 2:} The tandem bicycle is not a car but a vehicle designed for two riders.
\item \textbf{Premise 3:} The small metallic minivan is a car and there is only one such vehicle in the image.
\end{itemize}

\begin{tabular}{lp{12cm}}
\textbf{Relevance:} &These premises are relevant as they clarify what counts as a car. Relevance is high (3).\\\\

\textbf{Acceptability:} & It's logical to distinguish between cars and bicycles. However, the description mentions a motorcycle, which might affect the count. Acceptability is medium (2).\\\\

\textbf{Credibility:} & The premises are based on the image, which includes the tandem bicycle. Credibility is high (3).\\\\

\textbf{Consistency:} & Debater B's image description includes the tandem bicycle, SUVs, and a motorcycle, aligning with the premises. Consistency is high (3).\\
\end{tabular}

\medskip 

After evaluating, only Debater B's premises have high scores in all aspects. The image description from B includes the tandem bicycle, which is gray but not a car. The small metallic minivan is the only car. Therefore, the number of gray cars (1) is equal to the number of minivans (1), so the answer is (B) No.

\medskip 
The correct answer is determined by evaluating the premises and image descriptions. Debater B's premises are consistent with their image description, correctly identifying the tandem bicycle as not a car. Thus, the number of gray cars equals the number of minivans.

Answer: B
\end{tcolorbox}

\begin{tcolorbox}[
  enhanced,
  colback=white, % Light gray background (10% gray)
  colframe=blue1, % Darker gray frame (50% gray)
  arc=3mm, % Rounded corners
  boxrule=1pt, % Frame thickness
  fonttitle=\bfseries, % Bold title
  halign title=flush center, 
  title=Example Trace (for Question and Image in Figure~1), % Optional title for the box % Full width
  breakable, % Allow page breaks within the box
  before skip=10pt, % Space before the box
  after skip=10pt % Space after the box
]
\small 
Yes, the blue umbrella is under the black umbrella. The image shows a boat with passengers holding various umbrellas, including a black and a blue one. The black umbrella is positioned at the front of the boat, while the blue umbrella is directly below it. Due to the angle and positioning, the blue umbrella appears partially obscured by the black umbrella. This arrangement indicates that the blue umbrella is indeed under the black umbrella.
\end{tcolorbox}
\newpage 

\section{Dataset Examples}
\label{app:examples}
% \appendix
\begin{tcolorbox}[
title = MathVista,
colback=yellow!20,
colbacktitle=yellow!70!black,
]
\begin{tabular}{cl}
\includegraphics[scale=.8]{mathvista3.jpg} & \raisebox{4cm}[0pt]{Is the derivative of the function positive between [1, 2]}\\
& \raisebox{4cm}[0pt]{assuming that it's differentiable?} \\
& \raisebox{3cm}[0pt]{Choices: Yes/No} \\
& \raisebox{2cm}[0pt]{Answer: Yes} \\
\end{tabular}
\vspace{-1.4cm}
\end{tcolorbox}

\begin{tcolorbox}[
title = MathVista,
colback=yellow!20,
colbacktitle=yellow!70!black,
]
\begin{tabular}{cl}
\includegraphics[scale=.9]{mathvista1.jpg} & \raisebox{4cm}[0pt]{Which region is larger? R1 or R2? A. R1 B. R2?} \\
& \raisebox{3cm}[0pt]{Choices: R1\hspace{1em} R2\hspace{1em} R5\hspace{1em} R3\hspace{1em} R4} \\
& \raisebox{2cm}[0pt]{Answer: R2} \\
\end{tabular}
\vspace{-1cm}
\end{tcolorbox}

\begin{tcolorbox}[
title = MME,
colback=yellow!20,
colbacktitle=yellow!70!black,
]
\begin{tabular}{cl}
\includegraphics[scale=.4]{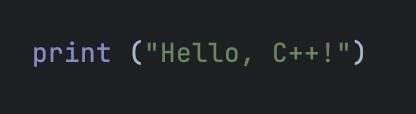} & \raisebox{3.5em}[0pt]{Is a python code shown in the picture?}\\
& \raisebox{3em}[0pt]{Please answer yes or no.} \\
& \raisebox{2.5em}[0pt]{Choices: Yes/No} \\
& \raisebox{2em}[0pt]{Answer: Yes} \\
\end{tabular}
\vspace{-1cm}
\end{tcolorbox}

\begin{tcolorbox}[
title = MME,
colback=yellow!20,
colbacktitle=yellow!70!black,
]
\begin{tabular}{cl}
\includegraphics[scale=.20]{mme_ex_chalet.jpg} & \raisebox{3cm}[0pt]{Does this image describe a place of tower?}\\\\
& \raisebox{3cm}[0pt]{Please answer yes or no.}\\  \\
& \raisebox{3cm}[0pt]{Choices: Yes/No} \\
& \raisebox{2.6cm}[0pt]{Answer: No} \\
\end{tabular}
\vspace{-2.3cm}
\end{tcolorbox}

\begin{tcolorbox}[
title = MMMU,
colback=yellow!20,
colbacktitle=yellow!70!black,
]
\begin{tabular}{cl}
\includegraphics[scale=.4]{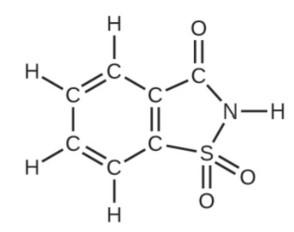} & \raisebox{3cm}[0pt]{How many molecules of the sweetener saccharin can be prepared}\\
& \raisebox{3cm}[0pt]{from 30~$C$ atoms, 25~$H$ atoms, 12~$O$ atoms, 8 $S$~atoms, and 14}\\  \\
& \raisebox{3.5cm}[0pt]{$N$~atoms assuming that it's differentiable?} \\
& \raisebox{3cm}[0pt]{Choices: 3\hspace{1em} 4\hspace{1em} 5\hspace{1em} 6} \\
& \raisebox{2.6cm}[0pt]{Answer: ~4} \\
\end{tabular}
\vspace{-2.5cm}
\end{tcolorbox}

\begin{tcolorbox}[
title = MMMU,
colback=yellow!20,
colbacktitle=yellow!70!black,
]
\begin{tabular}{cll}
\includegraphics[scale=.2]{mmmu_painting.png} & \multicolumn{2}{l}{\raisebox{3.5cm}[0pt]{What is NOT exhibited in the painting?}}\\ \\
& \raisebox{3.5cm}[0pt]{Choices:} &  \raisebox{3.5cm}[0pt]{A: hierarchical scale} \\ 
&  &\raisebox{3.5cm}[0pt]{B: graphic representation of horror and despair} \\ 
& & \raisebox{3.5cm}[0pt]{C: a sense of immediacy and drama} \\ 
& & \raisebox{3.5cm}[0pt]{D: use of sharply contrasting light and shade}
\\
& \raisebox{3.1cm}[0pt]{Answer:} &  \raisebox{3.1cm}[0pt]{A} \\
\end{tabular}
\vspace{-3.0cm}
\end{tcolorbox}

\end{document}